\documentclass{article}
\usepackage{colm2024_conference}

\usepackage{microtype}

\definecolor{openmossblue}{HTML}{223E9A}
\usepackage[colorlinks=true,
            linkcolor=openmossblue,
            urlcolor=openmossblue,
            citecolor=openmossblue,
            filecolor=openmossblue]{hyperref}
\usepackage{url}
\usepackage{booktabs}
\usepackage{graphicx}
\usepackage{setspace}
\usepackage{enumitem}

\usepackage{times}
\usepackage{latexsym}
\usepackage{amsmath}

\usepackage{xcolor}
\usepackage[dvipsnames]{xcolor}
\usepackage[table]{xcolor}
\usepackage{amssymb}
\usepackage{bm}
\usepackage{wrapfig}
\usepackage{stackengine}
\usepackage{multirow}
\usepackage{soul}
\usepackage{colortbl}
\usepackage{changepage,threeparttable}

\definecolor{deltaBg}{RGB}{220,230,255}
\newcommand{\rowhighlight}{\rowcolor{deltaBg}}
\newcommand{\normalrow}{\rowcolor{gray!15}}

\usepackage[T1]{fontenc}

\usepackage[utf8]{inputenc}

\usepackage{microtype}

\usepackage{inconsolata}
\usepackage{fancyhdr}

\usepackage{graphicx}
\usepackage{subcaption}
\usepackage{multirow}
\usepackage{tcolorbox}

\newcommand
\blfootnote[1]{%
\begingroup 
\renewcommand\thefootnote{}%
\footnote{#1}%
\addtocounter{footnote}{-1}%
\endgroup 
}

\title{SRPO: Self-Referential Policy Optimization for Vision-Language-Action Models}

\colmfinalcopy

\author{
Senyu Fei$^{2,3,*}$ \hspace{.2em}
Siyin Wang$^{1,3,*}$ \hspace{.2em}
Li Ji$^{1,3}$ \hspace{.1em}
Ao Li$^{3}$ \hspace{.1em}
Shiduo Zhang$^{1}$ \hspace{.1em}
Liming Liu$^{3}$ \hspace{.1em}
\\
\textbf{
Jinlong Hou$^{3}$ \hspace{.1em}
Jingjing Gong$^{3,\dagger}$ \hspace{.1em}
Xianzhong Zhao$^{2}$ \hspace{.2em}
Xipeng Qiu$^{1,3, \dagger}$
}
\\
[1ex]
\texttt{feisenyu@outlook.com,siyinwang20@fudan.edu.cn} \\
[1ex]
$^{1}$Fudan University \hspace{.1em}
$^{2}$Tongji University  \hspace{.1em}
$^{3}$Shanghai Innovation Institute \\
}

\fancypagestyle{firstpage}{
  \lhead{\begin{picture}(0,0)\put(0,-6){\includegraphics[width=0.3\linewidth]{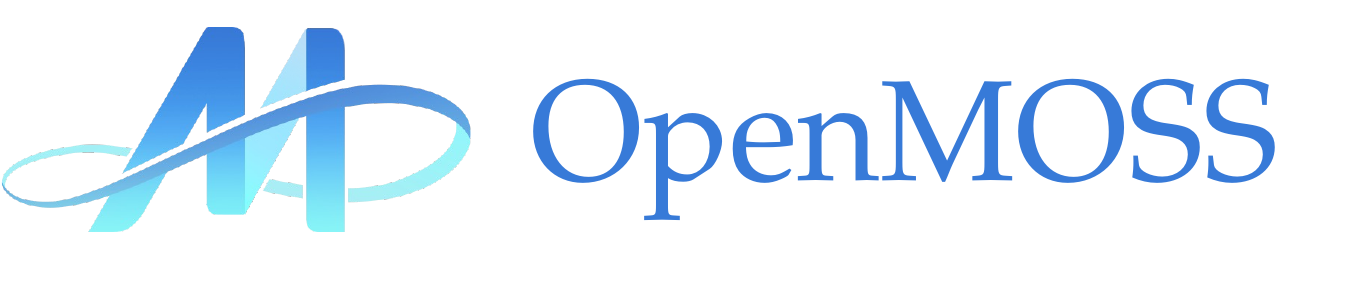}}\end{picture}}
}

\begin{document}

\maketitle
\blfootnote{\hspace{-0.67em}\textsuperscript{*}Equal Contribution.}
\blfootnote{\hspace{-0.67em}\textsuperscript{$\dagger$}Corresponding Authors.}

\thispagestyle{firstpage}

\begin{figure}[htbp]
    \centering
    \vspace{-7mm}
    \includegraphics[width=1.0\linewidth]{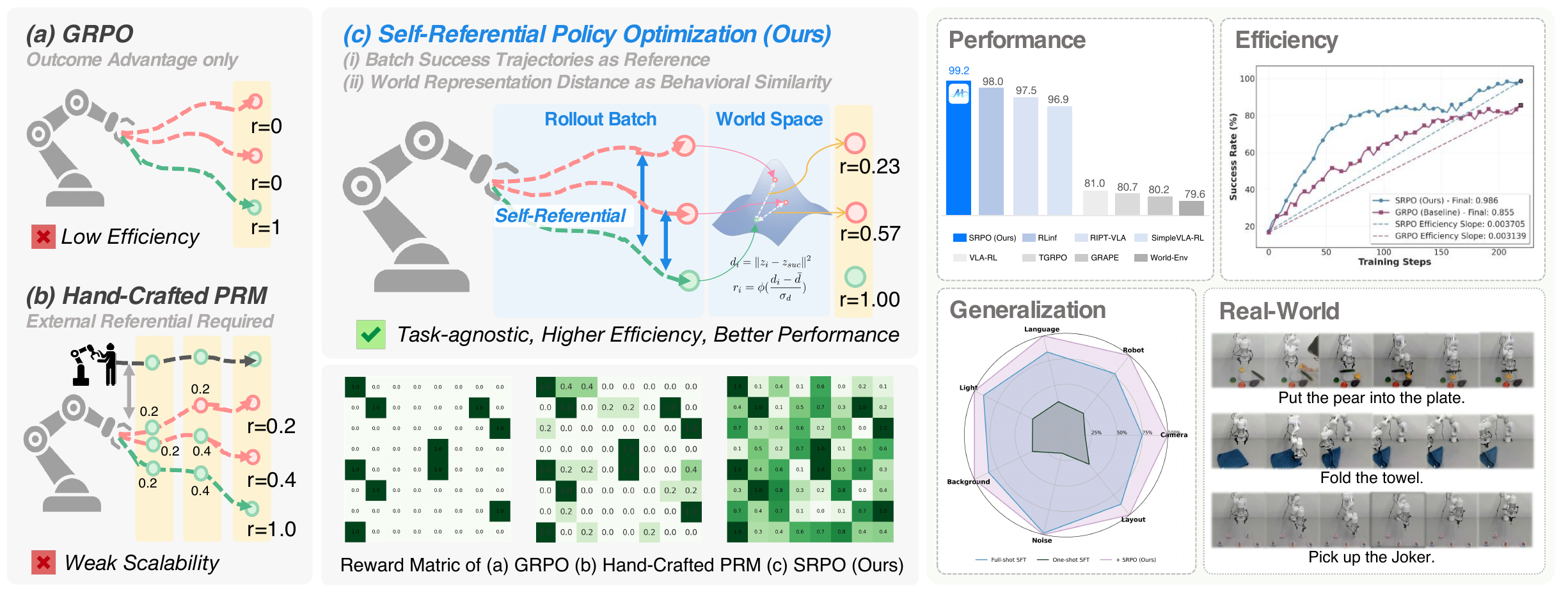}
    \caption{Overview of Self-Referential Policy Optimization (SRPO). Existing approaches for Vision-Language-Action (VLA) reinforcement learning face significant limitations: (a) methods like GRPO rely solely on sparse outcome rewards, providing limited learning signal, while (b) hand-crafted process reward modeling (PRM) requires costly external demonstrations and task-specific engineering. In contrast, our SRPO framework introduces a self-referential paradigm that leverages (i) in-batch successful trajectories and (ii) latent world representations to construct progress-wise rewards, enabling efficient utilization of failure trajectories. Extensive experimental evaluation demonstrates that SRPO achieves (1) state-of-the-art performance, (2) superior training efficiency, (3) stronger generalization capabilities, and (4) improved real-world performance.}
    \label{fig:placeholder}
\end{figure}

\begin{abstract}
Vision-Language-Action (VLA) models excel in robotic manipulation but are constrained by their heavy reliance on expert demonstrations, leading to demonstration bias and limiting performance. Reinforcement learning (RL) is a vital post-training strategy to overcome these limits, yet current VLA-RL methods, including group-based optimization approaches, are crippled by severe reward sparsity. Relying on binary success indicators wastes valuable information in failed trajectories, resulting in low training efficiency. To solve this, we propose Self-Referential Policy Optimization (SRPO), a novel VLA-RL framework. SRPO eliminates the need for external demonstrations or manual reward engineering by leveraging the model's own successful trajectories, generated within the current training batch, as a self-reference. This allows us to assign a progress-wise reward to failed attempts.
A core innovation is the use of latent world representations to measure behavioral progress robustly. Instead of relying on raw pixels or requiring domain-specific fine-tuning, we utilize the compressed, transferable encodings from a world model's latent space. These representations naturally capture progress patterns across environments, enabling accurate, generalized trajectory comparison.
Empirical evaluations on the LIBERO benchmark demonstrate SRPO's efficiency and effectiveness. Starting from a supervised baseline with 48.9\% success, SRPO achieves a new state-of-the-art success rate of 99.2\% in just 200 RL steps, representing a 103\% relative improvement without any extra supervision. Furthermore, SRPO shows substantial robustness, achieving a 167\% performance improvement on the LIBERO-Plus benchmark.
\end{abstract}

\section{Introduction}

Vision-Language-Action (VLA) models~\citep{rt2,octo,openvla,pi0} have demonstrated remarkable capabilities in robotic manipulation by leveraging large-scale pre-trained vision-language models. However, existing VLA systems rely heavily on expert demonstrations and tend to overfit small downstream datasets, resulting in a strong demonstration bias that prevents them from surpassing human performance~\citep{libero-plus, zhang2025vlabench}. 
To address such limitations, many recent studies have shown that reinforcement learning (RL) is an effective post-training strategy that can substantially improve VLA performance, both in-distribution and out-of-distribution~\citep{guo2025improving,rl4vla,vla-rl}. Among RL methods, group-based optimization approaches such as GRPO~\citep{shao2024deepseekmath} have become particularly attractive, providing a simple yet highly effective learning paradigm for VLA fine-tuning~\citep{simplevla-rl,zang2025rlinf,ript-vla}.

Despite the remarkable progress in GRPO, it still suffers from sparsity challenges in reward signals \citep{chan2024dense,mu2024drs,cao2024enhancing}. This issue becomes particularly pronounced in VLA domains, as the computational cost of multi-turn trajectory rollouts is substantially higher and the inadequate utilization of information from failed trajectories significantly reduces training efficiency.
While recent work has explored process supervision to provide denser feedback, these approaches typically depend on expert demonstrations or hand-crafted task decompositions to define intermediate progress milestones \citep{vla-rl,tgrpo}, creating scalability limitations that contradict the goal of autonomous learning.

To address the reward sparsity challenge, we propose self-referential learning, where the model's own successful trajectories serve as reference standards to evaluate and guide failed attempts. While GRPO leverages outcome-only rewards for advantage estimation, our approach utilizes the entire trajectory batch more efficiently. This paradigm transforms the supervision problem from ``how do we obtain expert labels'' to ``how do we extract progress-wise reward from our own successes''.

A central challenge in this paradigm lies in quantitatively measuring the behavioral similarity between successful and failed trajectories to assess progress towards task completion.
Since trajectories are represented solely by visual observations, we require a reward model capable of extracting global state information to measure progress. While world models offer a natural approach for modeling such dynamics \citep{assran2025v,ali2025world}, traditional pixel-level world models suffer from poor generalization across domains or typically require extensive task-specific fine-tuning \citep{world-env}. We address this limitation by leveraging the latent representations learned by world models, which we find naturally capture transferable behavioral progress patterns across different environments. These latent world representations enable robust trajectory comparison without requiring precise environmental reconstruction or domain-specific training.

Building on these insights, we propose Self-Referential Policy Optimization (SRPO), which seamlessly integrates self-referential learning into the RL training structure. During each training iteration, we identify successful trajectories within the batch and use them as behavioral references through world model latent representations, while failed trajectories receive progress-wise rewards based on their behavioral alignment with successful patterns. We opt for trajectory-level rewards over fine-grained reward shaping as overly detailed hand-crafted dense signals can lead the policy to converge toward suboptimal solutions \citep{sutton2019bitterlesson,levine2024sporks}.

Empirical evaluation on the LIBERO \citep{liu2023libero} benchmark demonstrates the transformative impact of our approach. 
Starting from a supervised fine-tuned baseline with just one demonstration per task and a 48.9\% success rate, SRPO reaches 99.2\% success in merely 200 reinforcement learning steps, representing 103\% relative improvement that establishes new state-of-the-art performance. 
Importantly, this RL improvement requires no additional expert demonstrations or manual reward engineering. Our method also shows substantial improvements in robustness on the LIBERO-Plus \citep{libero-plus} benchmark, with 167\% performance improvement. Our proposed reward demonstrates its effectiveness in both simulation and real-world experiments.

Our contributions are threefold as follow:
\begin{enumerate}
    \item We propose SRPO, a novel VLA reinforcement learning framework that mitigates reward sparsity by using model-generated successful trajectories to provide progress-wise rewards for failed attempts, eliminating reliance on expert demonstration or task-specific engineering.
    \item We introduce a latent world representation-based progress-wise reward method that overcomes the generalization limitations and domain-specific training requirements of traditional pixel-level world models.
    \item Experimental results demonstrate that our method achieves state-of-the-art performance on the LIBERO benchmark and exhibits strong generalization capabilities on LIBERO-Plus, while eliminating the need for additional supervision during RL training, establishing a new paradigm for autonomous VLA learning.
\end{enumerate}

\section{Related Work}
\subsection{Vision-Language-Action Models}

The remarkable success of models trained on vast internet-scale datasets has demonstrated the power of scaling laws, motivating the field to increasingly focus on Vision-Language-Action (VLA) models for transferring these capabilities to embodied agents \citep{rt2, openvla, openvla-oft, pi0, pi0-fast,RoboOmni}. Due to the limited generalization and domain adaptability of supervised post-training, the field is increasingly turning to Reinforcement Learning (RL) to encourage exploration and enhance overall robustness  \citep{vla-rl, simplevla-rl, rl4vla}. These studies have explored the distinct characteristics of RL compared to SFT, employing RL algorithms to train both autoregressive \citep{ript-vla} and diffusion-based VLAs \citep{chen2025pi_}, while also developing several integrated training-and-inference frameworks. However, these approaches often suffer from sparse reward signals, which prevent effective utilization of information from failure trajectories and limit training efficiency. To address this, we propose a novel VLA-RL framework that leverages dense, zero-shot transferable rewards to enable highly efficient policy learning.

\subsection{Reinforcement Learning}

The application of Reinforcement Learning (RL) to Large Language Models (LLMs) has evolved significantly \citep{jaech2024openai,yang2025qwen3,guo2025deepseek,team2025kimi,comanici2025gemini,Fei2025UnleashingET}. RLHF \citep{bai2022training} successfully aligned models with human preferences using algorithms like PPO \citep{schulman2017proximal}. RLVR \citep{wen2025reinforcement} replacing expensive human preferences with automatically verifiable rewards. GRPO \citep{shao2024deepseekmath} bypasses the need for a learned reward model by using programmatic rewards. Recent works have also attempted to enhance stability and efficiency in optimization \citep{yu2025dapo,wang2025beyond,yang2025dcpo}. Since task success serves as a naturally verifiable reward, numerous works in the VLA domain have attempted to optimize policies using GRPO with binary (0/1) rewards \citep{simplevla-rl, zang2025rlinf, ript-vla, rl4vla}, while facing the challenge of sparse reward signals. As robotic trajectory rollouts are more time-consuming and resource-intensive, some works have begun to construct rewards using hand-crafted or task-specific priors \citep{tgrpo,vla-rl,shu2025rftf}. However, such methods requiring expert demonstrations or prior information are difficult to scale in online RL settings that encourage autonomous learning. To this end, we introduce a self-referential approach leveraging task-agnostic latent world representation to provide progress-wise rewards, thus fully utilizing failure trajectory information to enhance learning efficiency.

\section{Methods}
\subsection{Problem Formulation and Key Components}

\begin{figure*}
    \centering
    \includegraphics[width=1\linewidth]{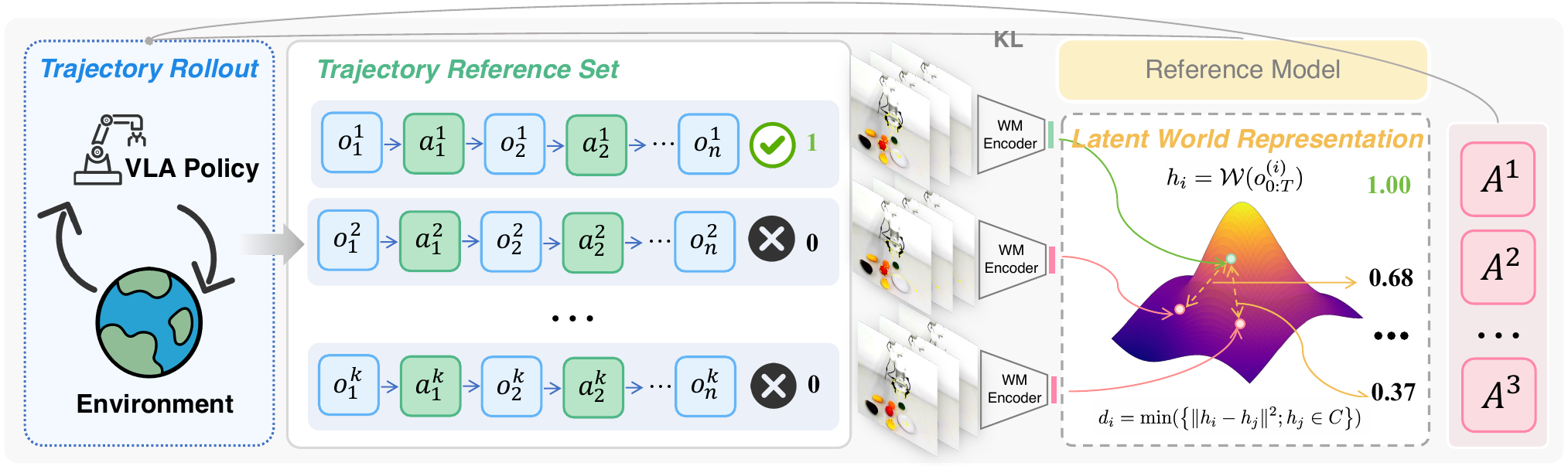}
    \caption{Overview of the SRPO method. During policy rollout, both successful and failed trajectories are collected in the Rollout Reference Set. For each trajectory, we employ a world model pre-trained on large-scale robotics video data \citep{assran2025v} as an encoder to extract latent world representations. Behavioral similarity is modeled as the L2 distance between trajectory embeddings in this space to yield progress-wise rewards. These rewards are subsequently used for advantage estimation and policy optimization under KL regularization.}
    \label{fig:placeholder}
\end{figure*}

Given the current observation $o_t$ at time step $t$, and a goal description $l$, the agent takes an action $a_t$ according to a policy $\pi_\theta(a_t|o_t, l)$ parameterized by $\theta$ and conditioned on a goal description $l$. The agent then receives the next observation $o_{t+1}$ from the environment, so on and so forth, until finally receiving a terminal observation $o_T$. If the agent successfully completes the task specified by the goal description $l$, it receives a positive reward; otherwise, it receives reward depending on how close it is to completing the task. The objective is to learn a policy that maximizes the reward received over time for a variety of goals $\mathbf{l}$ and the initial state of the environment $\mathbf{z}$.

\textbf{The observation function} is defined as $o_t \leftarrow O(z_t)$, where $O$ is the observation function that maps the environment state $z_t$ to the partial observation $o_t$ received by the agent.

\textbf{Action generation} is modeled by the policy function $\pi_\theta(a_t|o_t, l)$ which is the $\theta$ parameterized probability distribution over actions given the current observation $o_t$, goal description $l$.

\textbf{The environment function} $z_{t+1} \sim E(\cdot | z_t, a_t)$ defines the stochastic transition from the current environment state $z_t$ to the next environment state $z_{t+1}$ given the action $a_t$ taken by the agent.

\textbf{Trajectory Rollout} is the process of generating a sequence of observations and actions over time, starting from an initial environment state $z_0$.  The trajectory $\{(z_0, o_0, a_0, z_1, o_1, a_1, \ldots, z_T, o_T)\}$ are generated by iteratively applying these functions:
\begin{align}
\begin{aligned}
o_t &= O(z_t), &
a_t &\sim \pi_\theta(\cdot |o_t, l), &
z_{t+1} &\sim E(\cdot | z_t, a_t)
\end{aligned}
\end{align}
Note that the state of the environment $z_t$ is not directly accessible to the agent; instead, the agent relies on the observation $o_t$ derived from $z_t$ through the observation function $O$.

\textbf{The Reward function} $R(z_{0:T}, l)$ evaluates how well the state trajectory is performing in achieving the goal specified by $l$. This reward is usually provided by the environment at the end of the episode, based on whether the goal described by $l$ has been achieved. However, these rewards could be provided sparsely as 0 and 1, making it difficult for the agent to learn effective policies through traditional reinforcement learning methods. To address this, we adapt a world model to shape rewards $\hat{R}(o_{0:T}, \mathcal{S})$ for the failed trajectories to provide more informative feedback during training, as described in detail in Section~\ref{sec:world-progress-reward}.

\subsection{World progress reward modeling}
\label{sec:world-progress-reward}

As have mentioned earlier, neither the states $z_{0:T}$ nor the reward function $R(z_{0:T}, l)$ are directly accessible to the agent. There is only the sparse terminal reward signal provided at the end of each episode. We propose a world model-based task-agnostic reward shaping mechanism to provide progress-wise signals for unsuccessful rollout trajectories. The reward can be shaped as $\hat{R}(o_{0:T}, \mathcal{S})$, where $\mathcal{S} = \left\{o^{(i)}_{0:T} ; R(z^{(i)}_{0:T}, l) = 1, ~\forall i\right\}$ is the observation set of successful trajectories. 

First, we encode the observations of the unsuccessful trajectory and the successful reference trajectories, with a world-model encoder $\mathcal{W}$,
\begin{align}
h_i = \mathcal{W}(o^{(i)}_{0:T}).
\label{math:begin}
\end{align}
We then cluster the representations of the successful trajectories using the DBSCAN algorithm \citep{ester1996density} to obtain a set of representative centers. The distance to the closest center representation is then calculated; a smaller distance indicates a larger reward.
\begin{align}
    C &= \text{DBSCAN}(\mathcal{S}) \\
    d_i &= \min(\left\{\|h_i - h_j \|^2 ; h_j \in C  \right\})
\end{align}
Finally, the reward is calculated with
\begin{align}
    g_i = \begin{cases}
    1.0 & \text{for succes trajectory} \\
    \phi(\frac{d_i - \bar{d}}{\sigma_d}) & \text{for failed trajectory}
    \end{cases},
\label{math:end}
\end{align}
where $\phi(\cdot)$ is an activation function transforming the output to the range $(0, 1)$, and $\bar{d}$ and $\sigma_d$ are the mean and standard deviation, respectively, of the calculated distance for all failed trajectories.

\subsection{Self-Referential Policy Optimization}

We introduce \textit{Self-Referential Policy Optimization} (SRPO), a novel policy optimization framework for VLA models that leverages latent world representations to enable \textbf{self-referential advantage estimation} on \textbf{trajectory level}.

Following GRPO, the probability ratio, advantage and regularization terms are
\begin{align}
\begin{aligned}
r_{i,t}(\theta) &= \frac{\pi_\theta(a^{(i)}_t|o^{(i)}_t, l)}{\pi_{\theta_{old}}(a^{(i)}_t|o^{(i)}_t, l)}, &
\hat{A}_i &= \frac{g_i - \mu_{g}}{\sigma_{g}}, &
\omega(\theta) &= \beta D_{\text{KL}}\left( \pi_\theta \parallel \pi_{ref} \right).
\end{aligned}
\end{align}
The clipped surrogate objective is
\begin{align}
    \mathcal{L}^{\text{CLIP}}_{t,i}(\theta) = \min(r_{i,t}(\theta)\hat{A}_i, \text{clip}(r_{i,t}(\theta), 1-\epsilon, 1+\epsilon)\hat{A}_i)
\end{align}
The overall $\text{SRPO}$ objective, which we aim to maximize, is given by the expectation of the $\mathcal{L}_{\text{CLIP}}(\theta)$ function. This expectation is computed over time steps $t$ and sample IDs $i$ within the current training group, with an additional regularization term applied to the parameters $\theta$.
\begin{align}
    \mathcal{L}_{\text{SRPO}}(\theta) &= \mathbb{E}_{t,i}\mathcal{L}^{\text{CLIP}}_{t,i}(\theta) + \omega(\theta),
\end{align}
for simplicity, the expectation over different tasks $l$ and initial state $z_0$ have been omitted. The group statistics are derived from world progress rewards as follows:
\begin{equation}
    \mu_{\hat{R}} = \frac{1}{M}\sum_{j=1}^M {\hat{R}}_j, \
    \sigma_{\hat{R}} = \sqrt{\frac{1}{M}\sum_{j=1}^M (\hat{R}_j - \mu_{\hat{R}})^2 + \epsilon}.
\end{equation}
This self-referential approach enables the policy to learn from relative performance within trajectory groups, where rewards $\hat{R}_j$ are computed through our world progress reward model rather than traditional task-specific rewards. The KL divergence $D_{\text{KL}}(\pi_\theta \parallel \pi_{\text{ref}})$ term maintains policy stability.

\section{Experiments}
Our experimental design addresses several fundamental research questions: \textbf{RQ1:} Can SRPO achieve state-of-the-art performance on standard benchmarks?  \textbf{RQ2:} Does the method possess strong generalization capabilities to adapt to unseen task scenarios? \textbf{RQ3:} Can the latent world modeling-based reward mechanism accurately assess task progress, and does it outperform both pixel-level approaches and conventional video encoding methods? \textbf{RQ4:} Can SRPO improve training efficiency? \textbf{RQ5:} Can SRPO motivate novel trajectory exploration? \textbf{RQ6:} Whether the reward shaping method can successfully transfer to real-world robotic systems without requiring domain-specific expert demonstrations?

\subsection{Experimental Setup}

\begin{table}[t]\centering
\caption{Performance comparison on LIBERO benchmark. We evaluate mainstream VLA foundation models and RL-based methods. OpenVLA* incorporates action chunking and parallel decoding on the basis of OpenVLA. Policy Input notation: T (Thirdview), I (Instrcution), P (Proprio), W (Wristimage), D (Depth). Our approach, built upon one-shot SFT, achieves state-of-the-art results on LIBERO benchmark, with $\uparrow$ indicating performance gains over the one-shot baseline.}\label{tab: main_table}
\resizebox{0.96\linewidth}{!}{
\begin{tabular}{lcccccc}\toprule
\multirow{2}{*}{\textbf{Model}} & \multirow{2}{*}{\textbf{Policy Input}} & \multicolumn{5}{c}{\textbf{LIBERO}} \\\cmidrule{3-7}
& & Spatial & Object & Goal & Long & Avg \\\midrule
OpenVLA \citep{openvla} &T+I &84.7 &88.4 &79.2 &53.7 &76.5 \\
Pi0+fast \citep{pi0-fast}  &T+W+P+I &96.4 &96.8 &88.6 &60.2 &85.5 \\
Pi0 \citep{pi0}   &T+W+P+I &96.8 &98.8 &95.8 &85.2 &94.2 \\
SmolVLA \citep{shukor2025smolvla}   &T+W+P+I &93.0 &94.0 &91.0 &77.0 &88.8 \\
WorldVLA \citep{cen2025worldvla}  &T+I &85.6 &89.0 &82.6 &59.0 &79.1 \\
NORA \citep{nora}  &T+I &92.2 &95.4 &89.4 &74.6 &87.9 \\
CoT-VLA \citep{zhao2025cot}  &T+I &87.5 &91.6 &87.6 &69.0 &81.1 \\
UniVLA \citep{univla}  &T+I &96.5 &96.8 &95.6 &92.0 &95.2 \\
TraceVLA \citep{tracevla}  &T+I &84.6 &85.2 &75.1 &54.1 &74.8 \\
MolmoAct \citep{lee2025molmoact}  &T+I &87.0 &95.4 &87.6 &77.2 &86.6 \\
ThinkAct \citep{huang2025thinkact}  &T+I &88.3 &91.4 &87.1 &70.9 &84.4 \\
GR00T N1 \citep{bjorck2025gr00t}  &T+I &94.4 &97.6 &93.0 &90.6 &93.9 \\
3D-CAVLA \citep{bhat20253d}  &T+W+P+D+I &98.2 &99.8 &98.2 &96.1 &98.1 \\
OpenVLA-OFT \citep{openvla-oft}  &T+W+P+I &96.2 &98.3 &96.2 &90.7 &95.3 \\
OpenVLA*-Full &T+I &91.6 &95.3 &90.6 &86.5 &91.0 \\
\midrule
TGRPO \citep{tgrpo}  &T+I &90.4 &92.2 &81.0 &59.2 &80.7 \\
GRAPE \citep{zhang2024grape}  &T+I &88.5 &92.1 &83.1 &57.2 &80.2 \\
VLA-RL \citep{vla-rl}  &T+I &90.2 &91.8 &82.2 &59.8 &81.0 \\
World-Env \citep{world-env}  &T+I &87.6 &86.6 &86.4 &57.8 &79.6 \\
SimpleVLA-RL \citep{simplevla-rl} &T+I &98.2 &98.7 &98.8 &91.7 &96.9 \\
RIPT-VLA \citep{ript-vla} &T+W+P+I &99.0 &98.6 &98.6 &93.8 &97.5 \\
RLinf \citep{zang2025rlinf}  &T+W+P+I &\textbf{99.4} &99.8 &98.8 &94.0 &98.0
\\\midrule
OpenVLA*-One &T+I &63.6 &54.9 &59.6 &17.3 &48.9 \\
\rowhighlight
+ Offline SRPO &T+I &92.5 &96.8 &92.0 &88.7 &92.5 \\
&&\textcolor{ForestGreen}{$\uparrow$28.9} &\textcolor{ForestGreen}{$\uparrow$41.9} &\textcolor{ForestGreen}{$\uparrow$32.4} &\textcolor{ForestGreen}{$\uparrow$71.4} &\textcolor{ForestGreen}{$\uparrow$43.6}\\
\rowhighlight
\textbf{+ Online SRPO}  &T+I &98.8 &\textbf{100.0} &\textbf{99.4} &\textbf{98.6} &\textbf{99.2} \\
&&\textcolor{ForestGreen}{$\uparrow$35.2} &\textcolor{ForestGreen}{$\uparrow$45.1} &\textcolor{ForestGreen}{$\uparrow$39.8} &\textcolor{ForestGreen}{$\uparrow$81.3} &\textcolor{ForestGreen}{$\uparrow$50.3}\\
\bottomrule
\end{tabular} }
\end{table}

Main experiments are conducted on the \textbf{LIBERO} \citep{liu2023libero} benchmark, specifically on Goal, Spatial, Object and Long suites, each of which contains 10 tasks for evaluating a distinct model capability. To evaluate generalization robustness, we employ the \textbf{LIBERO-Plus} \citep{libero-plus} benchmark that introduces seven dimensions of perturbations. 

\textbf{Models.} For simulation experiments, we utilize a modified version of OpenVLA enhanced with action chunking and parallel decoding, which we refer to as \textbf{OpenVLA*} in the following sections. OpenVLA*-One refers to base model with one-traj-per-task SFT, while OpenVLA*-Full refers to full-shot SFT. More details can be found in Appendix~\ref{exp:real-world-details}.

\textbf{Baseline Methods.} Our baseline selection includes: \textbf{SimpleVLA-RL} \citep{simplevla-rl}, \textbf{RIPT-VLA} \citep{ript-vla} and \textbf{RLinf} \citep{zang2025rlinf} trained with GRPO, \textbf{VLA-RL} \citep{vla-rl} with PPO, \textbf{GRAPE} \citep{zhang2024grape} with trajectory-wise DPO, \textbf{TGRPO} \citep{tgrpo} with task-specific progress reward, \textbf{World-Env} \citep{world-env} utilizing world model as simulator. We also provide some imitation learning methods \citep{openvla,pi0-fast,pi0,shukor2025smolvla,cen2025worldvla,nora,zhao2025cot,univla,tracevla,lee2025molmoact,huang2025thinkact,bjorck2025gr00t,bhat20253d,openvla-oft} as reference, although they have different information inputs or pre-training data.

\textbf{Implementation Details.} In the scope of this work, we consider the observation $o_t$ to be the third view image, and the $l$ to be a language description of the goal. The environment is a physics simulator that takes in the current state $z_t$ (including object positions, velocities, etc.) and the action $a_t$ (e.g., end-effector commands) to produce the next state $z_{t+1}$. 
The architecture of our policy is based on a pre-trained Visual-Language-Action model, which integrates visual inputs and language descriptions to inform the agent's actions.

We leverage SiiRL \citep{wang2025distflowfullydistributedrl} to develop our training framework. Our pipeline begins with supervised fine-tuning using a single demonstration per task from the official checkpoint, followed by our SRPO method for online reinforcement learning post-training. To obtain shared latent world representations, we employ a large-scale video-pretrained latent world model, V-JEPA 2 \citep{assran2025v}. Training details and compute reports are provided in the Appendix~\ref{libero_train_detail}.

\begin{table}[!tp]
\centering
\caption{Robustness Evaluation on LIBERO-Plus Benchmark. OpenVLA-OFT+ refers to the OpenVLA-OFT model that has been trained on the LIBERO-Plus dataset. Our method, SRPO, applied to a one-shot SFT policy, not only significantly outperforms its base model but also surpasses the full-shot SFT baseline in all 7 dimensions, demonstrating superior generalization capability. The $\uparrow$ indicates performance gains over the One-shot SFT base model.}
\label{tab: plus_table}
\resizebox{0.98\linewidth}{!}{
\begin{tabular}{lcccccccc}
\toprule
\textbf{Model} & \multicolumn{7}{c}{\textbf{Perturbation Dimensions}} & \textbf{Total} \\
\cmidrule(lr){2-8}
& Camera & Robot-Init & Language & Light & Background & Noise & Layout & \\
\midrule
\normalrow
\multicolumn{9}{c}{\textit{Zero-Shot}} \\
\midrule
Pi0 & 13.8 & 6.0 & 58.8 & 85.0 & 81.4 & 79.0 & 68.9 & 53.6 \\
Pi0+fast & 65.1 & 21.6 & 61.0 & 73.2 & 73.2 & 74.4 & 68.8 & 61.6 \\
UniVLA & 1.8 & 46.2 & 69.6 & 69.0 & 81.0 & 21.2 & 31.9 & 42.9 \\
WorldVLA & 0.1 & 27.9 & 41.6 & 43.7 & 17.1 & 10.9 & 38.0 & 25.0 \\
OpenVLA & 0.8 & 3.5 & 23.0 & 8.1 & 34.8 & 15.2 & 28.5 & 15.6 \\
OpenVLA-OFT & 56.4 & 31.9 & 79.5 & 88.7 & 93.3 & 75.8 & 74.2 & 69.6 \\
OpenVLA-OFT\_w & 10.4 & 38.7 & 70.5 & 76.8 & 93.6 & 49.9 & 69.9 & 55.8 \\
OpenVLA-OFT\_m & 55.6 & 21.7 & 81.0 & 92.7 & 91.0 & 78.6 & 68.7 & 67.9 \\
NORA-long & 2.2 & 37.0 & 65.1 & 45.7 & 58.6 & 12.8 & 62.1 & 39.0 \\
RIPT-VLA & 55.2 & 31.2 & 77.6 & 88.4 & 91.6 & 73.5 & 74.2 & 68.4 \\
\cmidrule(lr){1-9}
OpenVLA*-Full & 12.8 & 39.4 & 68.5 & 63.4 & 75.0 & 34.8 & 62.7 & 51.1 \\
OpenVLA*-One & 3.2 & 14.0 & 27.6 & 25.7 & 32.7 & 6.4 & 26.4 & 19.4 \\
\rowhighlight
+ Online SRPO & 17.1 & 51.0 & 81.8 & 70.4 & 88.9 & 35.3 & 72.4 & 59.6 \\
& \textcolor{ForestGreen}{$\uparrow$13.9} & \textcolor{ForestGreen}{$\uparrow$37.0} & \textcolor{ForestGreen}{$\uparrow$54.2} & \textcolor{ForestGreen}{$\uparrow$44.7} & \textcolor{ForestGreen}{$\uparrow$56.2} & \textcolor{ForestGreen}{$\uparrow$28.9} & \textcolor{ForestGreen}{$\uparrow$46.0} & \textcolor{ForestGreen}{$\uparrow$40.2} \\
\midrule
\normalrow
\multicolumn{9}{c}{\textit{With Augmented Data}} \\
\midrule
OpenVLA-OFT+ & 92.8 & 30.3 & 85.8 & 94.9 & 93.9 & 89.3 & 77.6 & 79.5 \\
\cmidrule(lr){1-9}
OpenVLA*-Full & 69.4 & 49.6 & 66.3 & 88.2 & 88.5 & 78.7 & 70.3 & 73.0 \\
OpenVLA*-One & 12.8 & 23.0 & 30.0 & 42.0 & 49.6 & 23.3 & 34.5 & 30.7 \\
\rowhighlight
+ Online SRPO & 83.4 & 62.0 & 73.6 & 97.2 & 97.7 & 85.7 & 75.2 & 82.1 \\
& \textcolor{ForestGreen}{$\uparrow$70.6} & \textcolor{ForestGreen}{$\uparrow$39.0} & \textcolor{ForestGreen}{$\uparrow$43.6} & \textcolor{ForestGreen}{$\uparrow$55.2} & \textcolor{ForestGreen}{$\uparrow$48.1} & \textcolor{ForestGreen}{$\uparrow$62.4} & \textcolor{ForestGreen}{$\uparrow$40.7} & \textcolor{ForestGreen}{$\uparrow$51.4} \\
\bottomrule
\end{tabular}}
\end{table}

\subsection{Main Results}
Main results are summarized in Table~\ref{tab: main_table}, demonstrating that our approach achieves state-of-the-art performance in terms of the overall average success rate across all four suites. The key findings are as follows:

\textbf{(1) Effectiveness of Online Post-Training:} 
While the initial one-shot SFT baseline shows modest performance, SRPO significantly improves upon it, validating the efficacy of our online post-training paradigm in enhancing policy performance through environment interaction.

\textbf{(2) Superiority in Reward Design:} Our method outperforms RL-based methods like SimpleVLA-RL and RLinf, which rely on sparse outcome rewards, highlighting the importance of process rewards that provide denser supervisory signals. Furthermore, it surpasses approaches using manually designed process rewards like TGRPO and demonstrates that self-reference-based rewards, derived from pre-trained world models, offer a more effective learning signal than task-specific alternatives that require heuristic stage partitioning and reward tuning.

\textbf{(3) SOTA Performance with Limited Inputs:} Despite using only third-person visual observations and language instructions as input, our method outperforms several baselines that utilize additional modalities such as multiple camera views, proprioception \citep{pi0,pi0-fast,shukor2025smolvla,openvla-oft} and 3D data \citep{bhat20253d}. This demonstrates that SRPO can achieve state-of-the-art performance relying solely on visual inputs, highlighting its effectiveness and practical applicability.

\subsection{Generalization Performance}

To assess SRPO's generalization, we validate two key aspects: (1) whether SRPO training in LIBERO environments improves generalization over SFT, and (2) whether SRPO is more robust to perturbations than training on the pre-collected LIBERO-Plus dataset. The following key findings are summarized from Table~\ref{tab: plus_table}: 

(1) \textbf{SRPO Overcomes the Diversity Limitation of One-Shot SFT.} 
While the one-shot SFT baseline exhibits weak generalization due to limited trajectory diversity, SRPO post-training enables it to surpass even the full-shot SFT policy. This is attributed to SRPO's online interaction, which explores a broader range of trajectories than those contained in the static full-shot dataset. A detailed analysis of this trajectory diversity advantage is provided in Section~\ref{exp:traj}. 

(2) \textbf{SRPO also Surpasses Full-Shot SFT in Perturbed Data Settings.} 
When trained in perturbed environments, our method not only exceeds the performance of the full-shot SFT baseline but also outperforms OpenVLA-OFT (utilizes additional wrist image and proprioception) trained with full-shot SFT. This result underscores that the trajectory diversity afforded by SRPO plays a major role in improving generalization, compensating for and exceeding the advantage of more complex input modalities.

\section{Analysis}

\subsection{Can Latent World Representation Improve Reward Shaping?}

We compare our approach against two alternative reward formulations: (1) pixel-level progress reward \citep{wen2025reinforcement}, which computes similarity based on pixel-wise features, and (2) ImageBind-based progress reward, derived from the general-purpose vision embeddings model ImageBind \citep{girdhar2023imagebind}.

\begin{wraptable}{r}{0.55\textwidth}
\vspace{-10pt}
\centering
\small
\caption{Progress Reward Benchmark Results. Our method achieved a better level than the baseline in all 5 indicators.}
\label{tab:progress_benchmark}
\begin{tabular}{lccccc}
\toprule
\textbf{Method} & \textbf{SC} & \textbf{Mono} & \textbf{MMD} & \textbf{JS} & \textbf{SMD}\\
\midrule
Pixel-level & 0.125 & 0.498 & 0.274 & 0.548 & 2.100 \\
ImageBind & 0.957 & 0.837 & 0.356 & 0.408 & 18.111 \\
SRPO (Ours) & \textbf{0.998} & \textbf{0.992} & \textbf{0.615} & \textbf{0.572} & \textbf{188.799} \\
\bottomrule
\end{tabular}
\vspace{-5pt}
\end{wraptable}

\begin{figure*}[tbp]
    \centering
    \includegraphics[width=\textwidth]{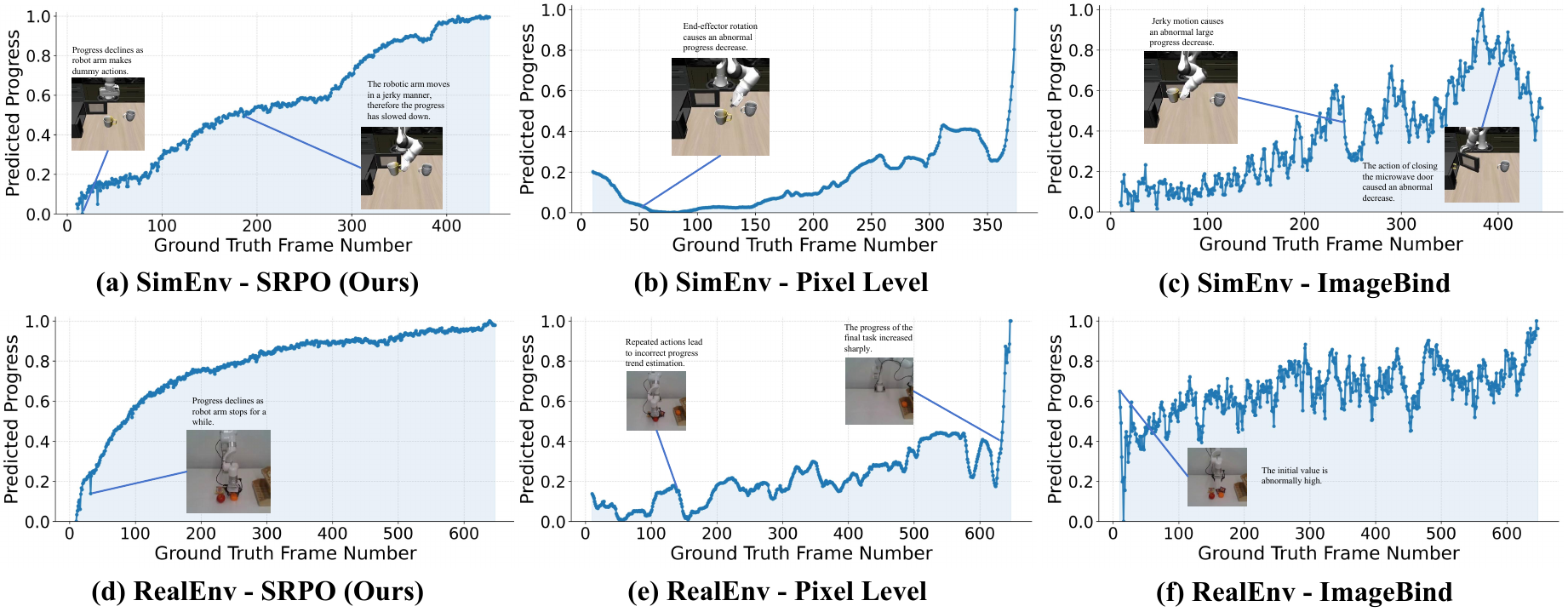}
    \caption{Comparison of progress estimation methods in simulated (a-c) and real-world (d-f) environments. Our SRPO reward (a,d) provides monotonic and physically plausible progress estimation. Pixel-level rewards (b,e) show sensitivity to perceptual changes, while ImageBind rewards (c,f) exhibit erratic trends from jerky motions.}
    \label{fig:progress_analysis}
\end{figure*}

\textbf{Qualitative Visualization.} 
We evaluate our progress reward approach on several tasks (e.g., placing a cup in a microwave, tidying a desk).
As shown in Figure~\ref{fig:progress_analysis}, our method produces smoother, more monotonic progress curves that better reflect task structure, especially for long-horizon tasks with repeated sub-tasks. In contrast, pixel-level and ImageBind-based methods generate fluctuating, non-monotonic progress signals. This improvement comes from embedding entire trajectories into a latent space that captures physical regularities, allowing for more accurate estimation of task progress.
Further reward curves and success/failure trajectory examples are provided in Appendix~\ref{app:reward_ana}. 

\textbf{Quantitative Comparison.} We carefully selected 700 success trajectories progressing steadily towards the goal and 300 failure trajectories from multiple tasks in both simulation and real-world environments. We establish five metrics for evaluating progress rewards: \textit{Spearman Correlation (SC)} and \textit{Monotonicity (Mono)} are computed exclusively on success trajectories to measure the positive correlation with frame numbers and the trajectory's monotonicity, respectively; \textit{Maximum Mean Discrepancy (MMD)}, \textit{JS Divergence (JS)}, and \textit{Standardized Mean Difference (SMD)} are statistical measures that quantify the distinguishability between success and failure trajectories. For all five metrics, a higher value indicates better performance. As shown in Table \ref{tab:progress_benchmark}, our method provides more reasonable progress rewards for intermediate failure states and better distinguishes between success and failure trajectories, thereby more effectively utilizing the failure trajectories. The detailed formulations of our bench are provided in Appendix~\ref{app:progress_benchmark}.

\begin{wrapfigure}{r}{0.49\linewidth}
    \vspace{-10pt}
    \centering
    \includegraphics[width=\linewidth]{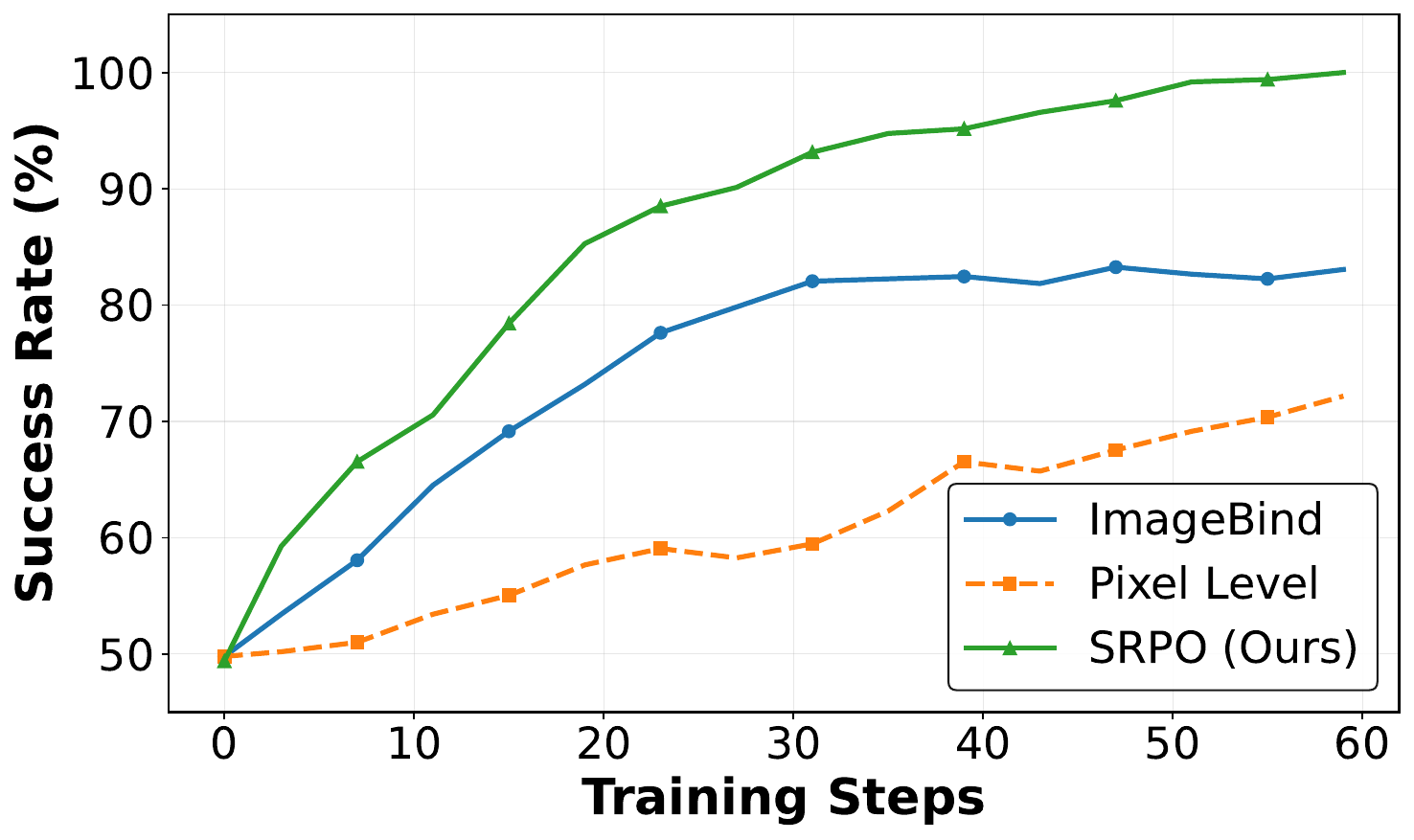}
    \caption{Training performance comparison using different progress reward formulations. Our SRPO-based reward enables stable and efficient learning, consistently outperforming both baselines.}
    \label{fig:training_comparison}
    \vspace{-10pt}
\end{wrapfigure}

\textbf{Training Effectiveness.} As shown in Figure~\ref{fig:training_comparison}, the pixel-level method exhibits notably slow convergence, suggesting that its reward signal provides insufficient guidance for effective policy optimization. While ImageBind demonstrates faster initial learning compared to the pixel-level approach, its performance plateaus around 85\% without further improvement, indicating that the suboptimal progress estimation hinders the model's ability to extract useful information from failure trajectories, ultimately causing training stagnation. We argue that the \textit{pixel-level} method is fundamentally limited because it relies solely on the final frame and exhibits sensitivity to minor pixel changes, and a general visual model like \textit{ImageBind} lacks the capability of understanding robotics-relevant physical concepts. Our experimental results confirm that effective progress estimation for VLAs requires rewards grounded in \textit{world progress} rather than simple perceptual similarity.

\subsection{Can SRPO Improve Training Efficiency?}

\begin{figure}[t]
    \centering
    \includegraphics[width=0.8\linewidth]{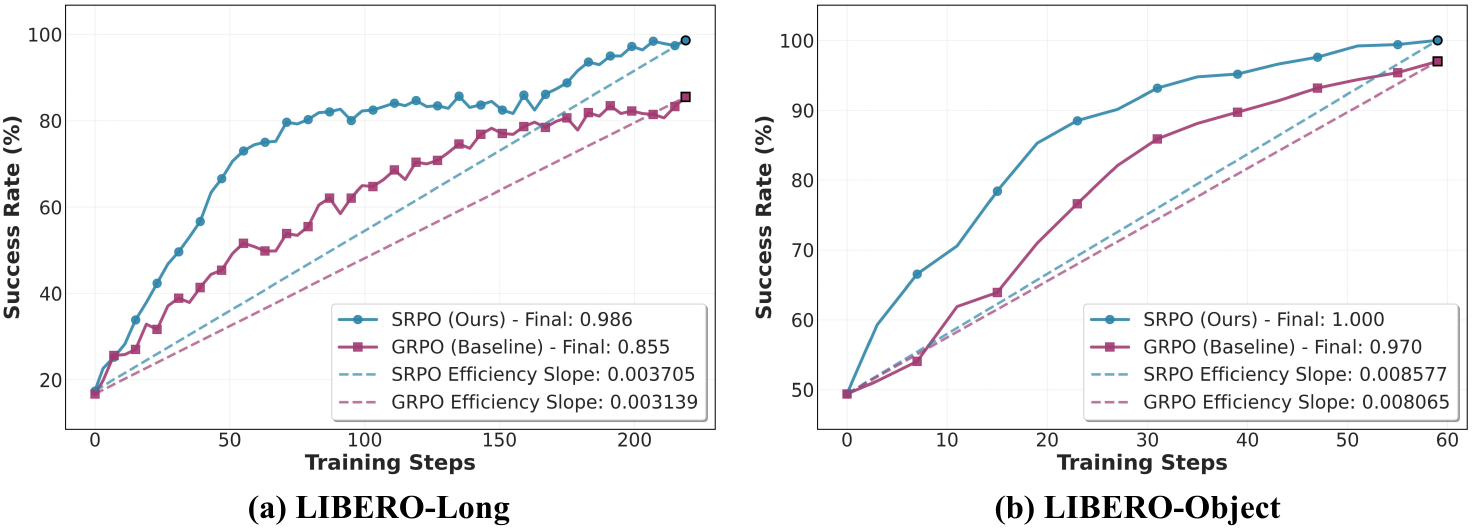}
    \caption{Training efficiency comparison between SRPO and GRPO: (a) LIBERO-Long, (b) LIBERO-Object.}
    \label{fig:efficiency_slope}
\end{figure}

Our results across the four suites are obtained using 79 steps (Spatial), 59 steps (Object), 103 steps (Goal), and 219 steps (Long), demonstrating significant advantages over SFT which requires tens of thousands of steps. Meanwhile, we compare the efficiency differences between SRPO and GRPO, as shown in Figure~\ref{fig:efficiency_slope}. The results clearly indicate that SRPO achieves a steeper efficiency slope than GRPO, especially for long-horizon tasks. Unlike GRPO, which essentially discards unsuccessful episodes, our self-referential framework can extract valuable learning signals from near-successful trajectories by recognizing and rewarding the productive segments within them, thereby guiding the policy more efficiently towards the final goal.

\subsection{Can SRPO Motivate Novel Trajectory Exploration?}
\label{exp:traj}

\begin{figure}[htbp]
    \centering
    \includegraphics[width=0.75\textwidth]{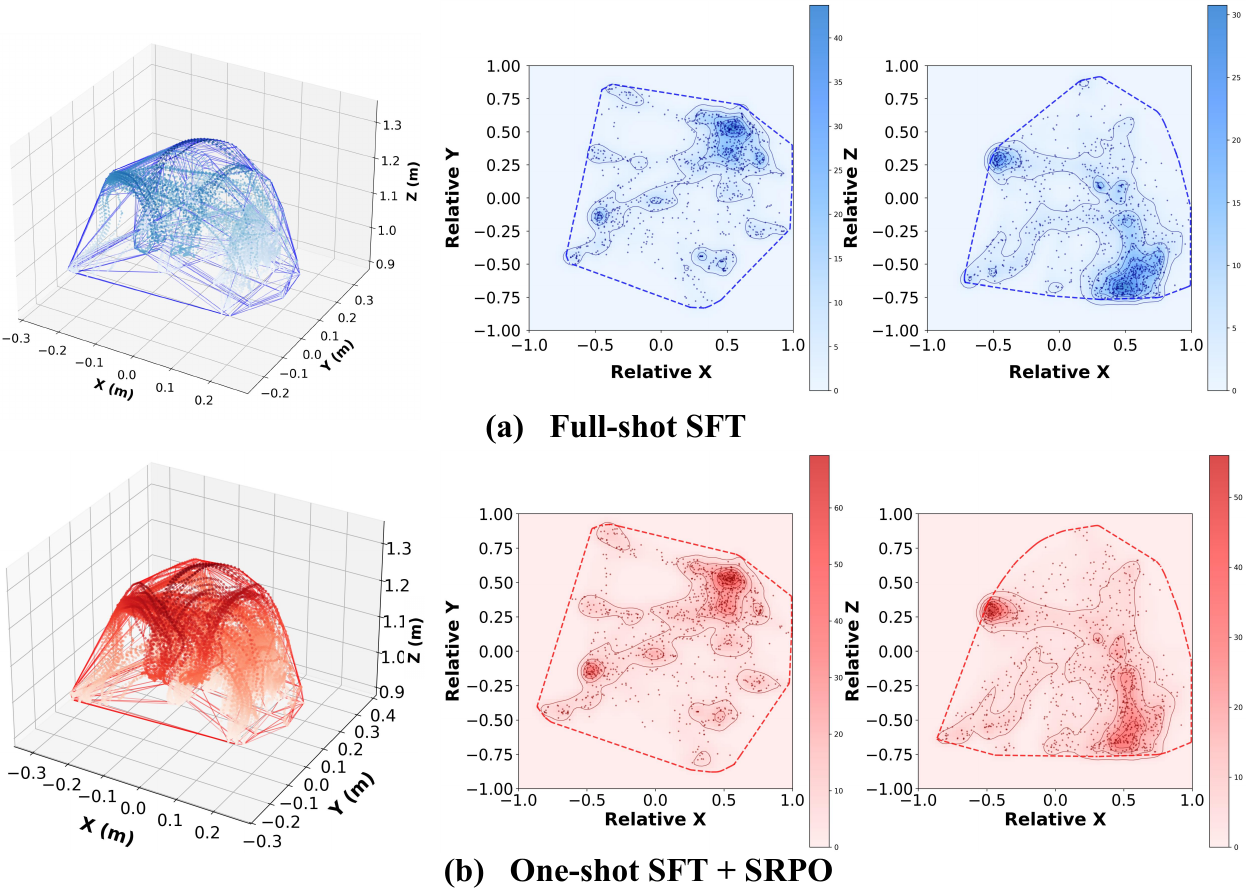}
    \caption{Action space comparison of end-effector trajectories between (a) full-shot supervised fine-tuning (SFT) and (b) SRPO online reinforcement learning (RL) policies.}
    \label{fig:action_diversity}
\end{figure}

A fundamental limitation of imitation learning is its inherent constraint to the state-action distribution present in the demonstration data. To quantitatively assess whether SRPO enables the policy to transcend these limitations and explore beyond the demonstrator's domain, we analyze the diversity of actions generated by the policy in \textit{LIBERO-Spatial} task suite. We performed rollouts on this suite using both the full-shot SFT weights and the RL fine-tuned weights, executing 10 trajectory episodes per task, with end-effector positions utilized as actions, which is visualize in Figure~\ref{fig:action_diversity}.

The results demonstrate that the online RL-trained policy exhibits two key advantages over the SFT baseline in terms of action distribution: (1) it explores previously unreachable regions; and (2) it generates more dispersed trajectories, indicating the model's propensity for spatial exploration rather than merely fitting to specific demonstration paths. 

\begin{wrapfigure}{r}{0.5\linewidth}
    \centering
    \includegraphics[width=\linewidth]{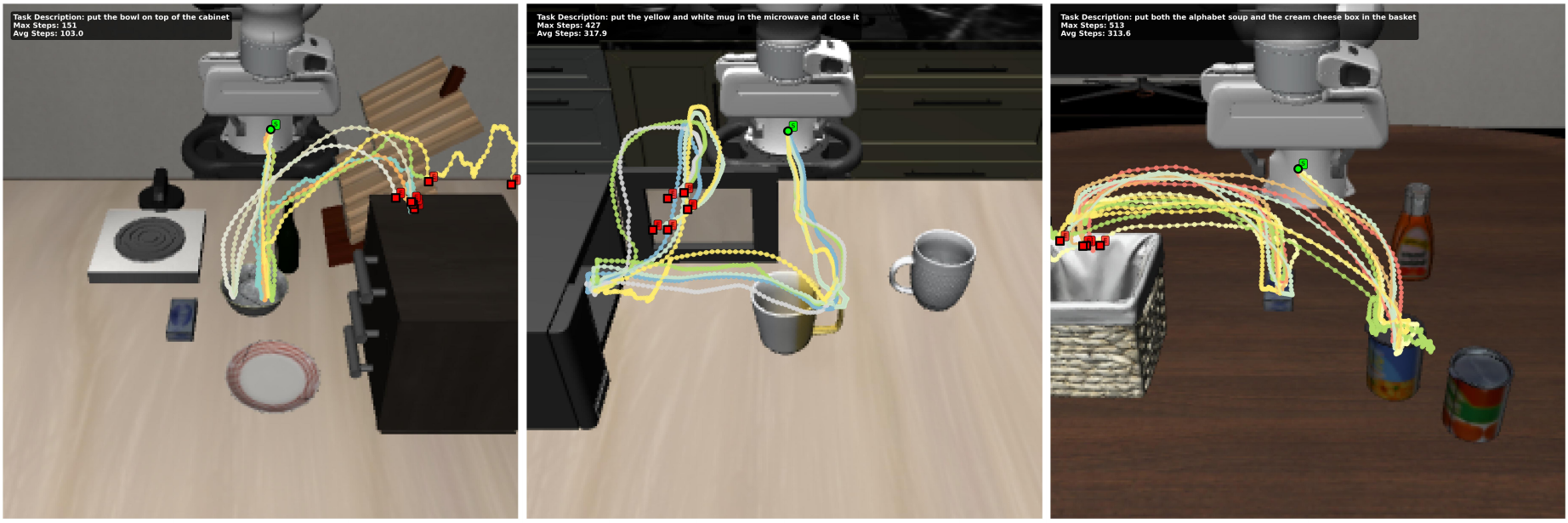}
    \caption{Visualization of end-effector trajectories across three tasks (from left to right): \textit{put the bowl on top of the cabinet}, \textit{put the yellow and white mug in the microwave and close it}, and \textit{put both the alphabet soup and the cream cheese box in the basket}.}
    \label{fig:traj}
\end{wrapfigure}

Furthermore, as illustrated in Figure~\ref{fig:traj}, even when the policy is initially exposed to only a single successful demonstration trajectory, the subsequent online RL fine-tuning enables it to discover novel strategies beyond the provided example. This exploration is evident not only in the diversity of spatial paths taken to approach the object but also in the variety of grasping positions it discovers.

This finding underscores a key strength of our method: the ability to autonomously acquire and leverage novel knowledge, in terms of both motor skills and affordance understanding, that was not present in the original, limited demonstration data, thereby significantly enhancing the policy's robustness and generalization.

\subsection{Can world progress rewarding modeling extend to real-world?}

\begin{wrapfigure}{r}{0.5\textwidth}
    \centering
    \includegraphics[width=\linewidth]{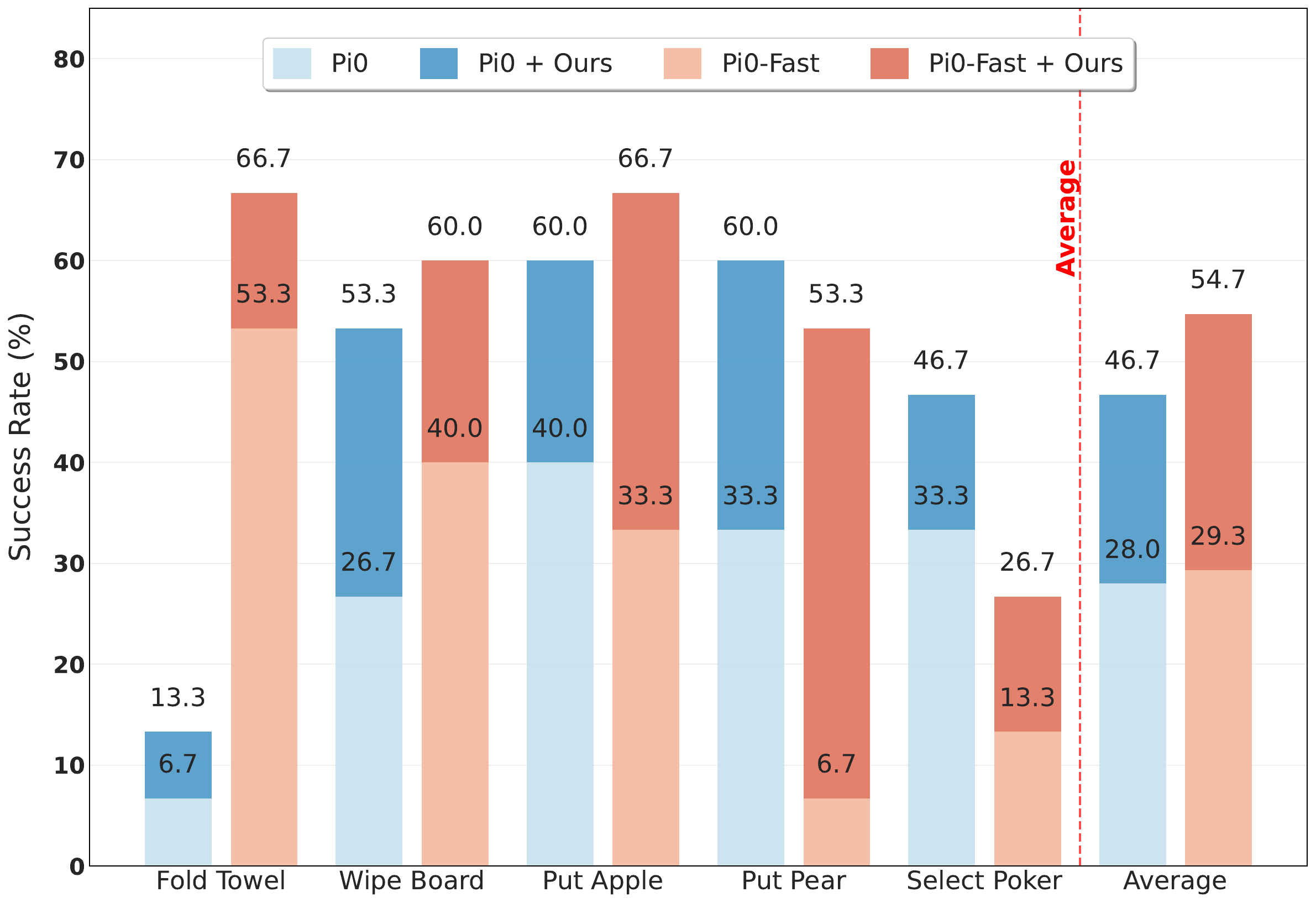}
    \caption{Real-world task success rates comparing supervised fine-tuning (SFT) baselines against our offline RL approach. Our method consistently improves performance across both diffusion-based ($\pi_0$) and autoregressive ($\pi_0$-FAST) VLA policies.}
    \label{fig:real_world_results}
    \vspace{-5mm}
\end{wrapfigure}

In this section, we investigate whether our proposed latent world progress rewarding method can effectively generalize to real-world robotic tasks. To this end, we conduct experiments with offline RL approach on five real-world manipulation tasks performed on an X-ARM 7 robot: \textit{Put apple into the plate}, \textit{Put pear into the plate}, \textit{Folding towels}, \textit{Cleaning whiteboard} and \textit{Select Poker}.

For each task, we compare two state-of-the-art vision-language-action (VLA) policy backbones: $\pi_0$ (a diffusion-based VLA model) and $\pi_0$-FAST (an autoregressive VLA enhanced with frequency-space tokenization). Both models are compared against standard supervised fine-tuning (SFT) baselines.

Both the diffusion-based $\pi_0$ and autoregressive $\pi_0$-FAST models demonstrate substantial performance improvements when enhanced with our reward shaping method, with average gains of $+66.8\%$ and $+86.7\%$, respectively. The most significant improvements are observed on tasks involving object placement and manipulation, underscoring the method's efficacy in adapting to perceptual variations. These consistent and substantial improvements across diverse tasks and policy architectures confirm that progress-aware reward modeling effectively transfers to real-world robotic manipulation, enabling more efficient policy optimization through better credit assignment and progression-aware weighting.

\section{Conclusion}

In this paper, we introduce Self-Referential Policy Optimization (SRPO), a novel VLA reinforcement learning approach leveraging task-agnostic latent world representation to provide progress-wise reward. By using self-referential learning, we eliminate reliance on expert demonstration or task-specific engineering, and develop an efficient VLA RL framework that can fully utilize the information provided by the failure trajectories, achieving significant improvements in both performance and efficiency compared to the existing methods. Further analysis demonstrates that our method improves reward shaping, motivates novel trajectory exploration, and the world progress reward modeling can be effectively applied to real-world robotic tasks. This work establishes a novel paradigm for efficient VLA reinforcement learning by demonstrating that pre-trained world model latent representations provide a powerful, generic substrate for progress assessment.

\newpage
\bibliography{colm2024_conference}
\bibliographystyle{colm2024_conference}

\clearpage
\setcounter{page}{1}
\maketitle

\appendix

\section{Progress Reward Benchmark}
\label{app:progress_benchmark}

Based on the carefully curated multi-task dataset of 700 human-annotated successful trajectories and 300 failure trajectories across diverse task domains, we propose a practical evaluation framework to assess the quality of progress reward functions. All successful trajectories are manually selected to exhibit approximately linear growth characteristics, meaning they demonstrate consistent forward progression without significant regressions or backtracking (e.g., no instances of dropping objects and having to retrieve them). To better test the model's generalization performance, the selected trajectories contain various perturbations including camera viewpoint changes, lighting variations, compounding objects, sensor noise and background distractions. The benchmark focuses on five core metrics that capture the essential properties of effective progress signals across multiple tasks and perturbation scenarios.

\subsection{Core Evaluation Metrics}

\begin{itemize}
    \item \textbf{Temporal Correlation}: Measures the correlation between progress values and frame numbers across all tasks using Spearman's rank correlation coefficient:
    \begin{equation}
        \rho = \frac{1}{N} \sum_{k=1}^{N} \frac{\sum_{i=1}^{T_k}(x_i^{(k)} - \bar{x}^{(k)})(y_i^{(k)} - \bar{y}^{(k)})}{\sqrt{\sum_{i=1}^{T_k}(x_i^{(k)} - \bar{x}^{(k)})^2 \sum_{i=1}^{T_k}(y_i^{(k)} - \bar{y}^{(k)})^2}},
    \end{equation}
    where $N$ is the number of tasks, $T_k$ is the trajectory length for task $k$, $x_i^{(k)}$ represents frame numbers and $y_i^{(k)}$ represents progress values. Higher absolute values indicate stronger monotonic relationships.

    \item \textbf{Temporal Monotonicity}: Quantifies the average percentage of steps where progress increases across all tasks:
    \begin{equation}
        M_{\text{mono}} = \frac{1}{N} \sum_{k=1}^{N} \frac{1}{T_k-1} \sum_{t=1}^{T_k-1} \mathbb{I}(r_{t+1}^{(k)} > r_t^{(k)}),
    \end{equation}
    where $\mathbb{I}$ is the indicator function and $r_t^{(k)}$ is the progress at step $t$ for task $k$. Values closer to 100 indicate stronger monotonic progression.

    \item \textbf{Distribution Separation (MMD)}: Evaluates the average separation between success and failure trajectories across tasks using Maximum Mean Discrepancy:
    \begin{equation}
        \text{MMD} = \frac{1}{N} \sum_{k=1}^{N} \left\| \frac{1}{n_k}\sum_{i=1}^{n_k}\phi(R_{s,k}^{(i)}) - \frac{1}{m_k}\sum_{j=1}^{m_k}\phi(R_{f,k}^{(j)}) \right\|_{\mathcal{H}}^2,
    \end{equation}
    where $R_{s,k}$ and $R_{f,k}$ represent final progress values for success and failure trajectories in task $k$, $n_k$ and $m_k$ are the respective sample sizes, and $\phi$ is the feature map in reproducing kernel Hilbert space $\mathcal{H}$.

    \item \textbf{Jensen-Shannon Divergence}: Measures the average distributional divergence between success and failure trajectories across tasks:
    \begin{equation}
        \text{JSD} = \frac{1}{N} \sum_{k=1}^{N} \left[ \frac{1}{2} D_{\text{KL}}(P_{\text{success}}^{(k)} \| M^{(k)}) + \frac{1}{2} D_{\text{KL}}(P_{\text{failure}}^{(k)} \| M^{(k)}) \right],
    \end{equation}
    where $M^{(k)} = \frac{1}{2}(P_{\text{success}}^{(k)} + P_{\text{failure}}^{(k)})$ is the mixture distribution for task $k$, and $D_{\text{KL}}$ denotes the Kullback-Leibler divergence.

    \item \textbf{Standardized Mean Difference}: Measures the average separation effect size between success and failure trajectories across tasks:
    \begin{equation}
        \text{SMD} = \frac{1}{N} \sum_{k=1}^{N} \frac{\mu_{\text{success}}^{(k)} - \mu_{\text{failure}}^{(k)}}{\sigma_{\text{pooled}}^{(k)}},
    \end{equation}
    where $\mu_{\text{success}}^{(k)}$ and $\mu_{\text{failure}}^{(k)}$ are the means for task $k$, and $\sigma_{\text{pooled}}^{(k)} = \sqrt{\frac{(n_k-1)(\sigma_{\text{success}}^{(k)})^2 + (m_k-1)(\sigma_{\text{failure}}^{(k)})^2}{n_k + m_k - 2}}$ is the pooled standard deviation for task $k$.
\end{itemize}

\begin{figure}[t]
    \centering
    \includegraphics[width=0.85\linewidth]{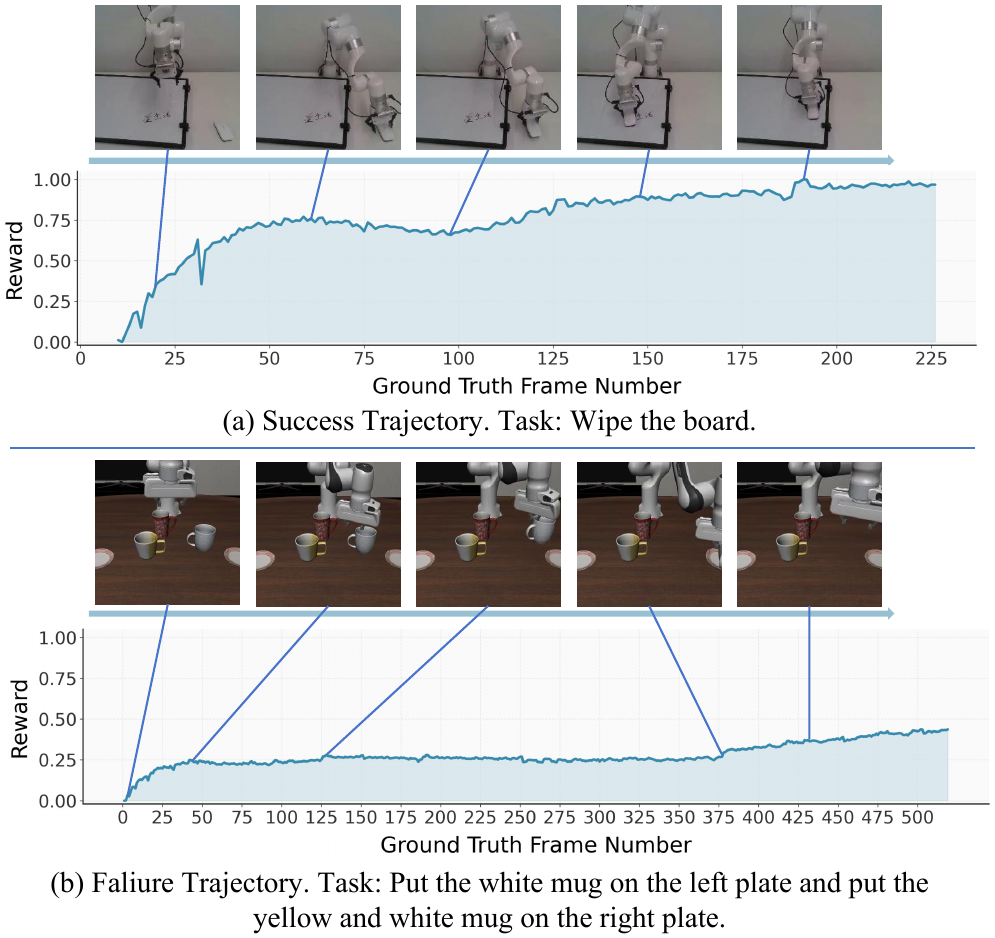}
    \caption{Visualization of reward signals between successful and failure trajectories. (a) A success trajectory (task: wipe the board) exhibits a reward curve that rises smoothly; a slight dip in reward corresponds to a pause during grasping. (b) A failure trajectory (task: put mugs on plates) shows a stagnant reward signal, failing to reach a high value, as the model fails to locate the second mug.}
    \label{fig:detail_curve}
\end{figure}

\subsection{Evaluating Progress Reward Quality}

A high-quality progress reward function should exhibit several key characteristics that can be assessed through the following complementary metrics:

\begin{itemize}
    \item \textbf{Temporal Consistency}: The progress values should demonstrate strong \textit{temporal correlation} ($\rho$) with monotonically increasing patterns (\textit{temporal monotonicity} $M_{\text{mono}}$). Effective progress rewards should show consistent improvement over time, with correlation values approaching 1.0 and monotonicity scores near 100\% indicating smooth, predictable progression toward task completion.

    \item \textbf{Distribution Discriminability}: Successful and failed trajectories should be well-separated in the progress reward space. This is quantified through multiple complementary measures:
    \begin{itemize}
        \item \textit{Maximum Mean Discrepancy (MMD)}: Captures distributional differences in reproducing kernel Hilbert spaces, with larger values indicating better separation between success and failure trajectories.
        \item \textit{Jensen-Shannon Divergence (JSD)}: Measures the information-theoretic divergence between success and failure distributions, where values closer to $\ln{2}$ suggest maximal discriminability.
        \item \textit{Standardized Mean Difference (SMD)}: Provides an effect size measure for the separation between success and failure means, with larger absolute values indicating stronger differentiation.
    \end{itemize}
\end{itemize}

The combination of these metrics provides a comprehensive framework for evaluating progress reward quality, emphasizing both the internal consistency within successful trajectories and the external discriminability between success and failure outcomes.

\section{Reward Analysis Details}
\label{app:reward_ana}

\textbf{Reward construction for success traj.} To ensure a fair and standardized comparison, we adopt the following normalized procedure for progress reward computation: For SRPO and ImageBind, we compute video embeddings using a cumulative sliding window approach: starting from frames 0-10, we progressively extend the window through frames 1-11, 1-12, and so forth, until the final window spanning frames 1 to the penultimate frame, which generates a sequence of embeddings representing cumulative visual context up to each time step. Then, we compute their L2 distances to the embedding of the entire video sequence, and normalize the progress reward by assigning a value of 0 to the frame with the maximum distance and 1 to the frame with the minimum distance. For the pixel-level method (RLVR), we calculate the L1 distance between each frame from frame 10 to the penultimate frame and the final frame, applying the same normalization scheme.

\textbf{Reward construction for failure traj.} For failure trajectories, we use the cluster centers of successful trajectories as reference points. We compute the minimum L2 distance from each failure trajectory segment to these success cluster centers, then normalize the progress reward by assigning a value of 0 to the segment with the maximum distance and 1 to the segment with zero distance (closest to success patterns).

\begin{figure*}[t]
    \centering
    \includegraphics[width=0.88\textwidth]{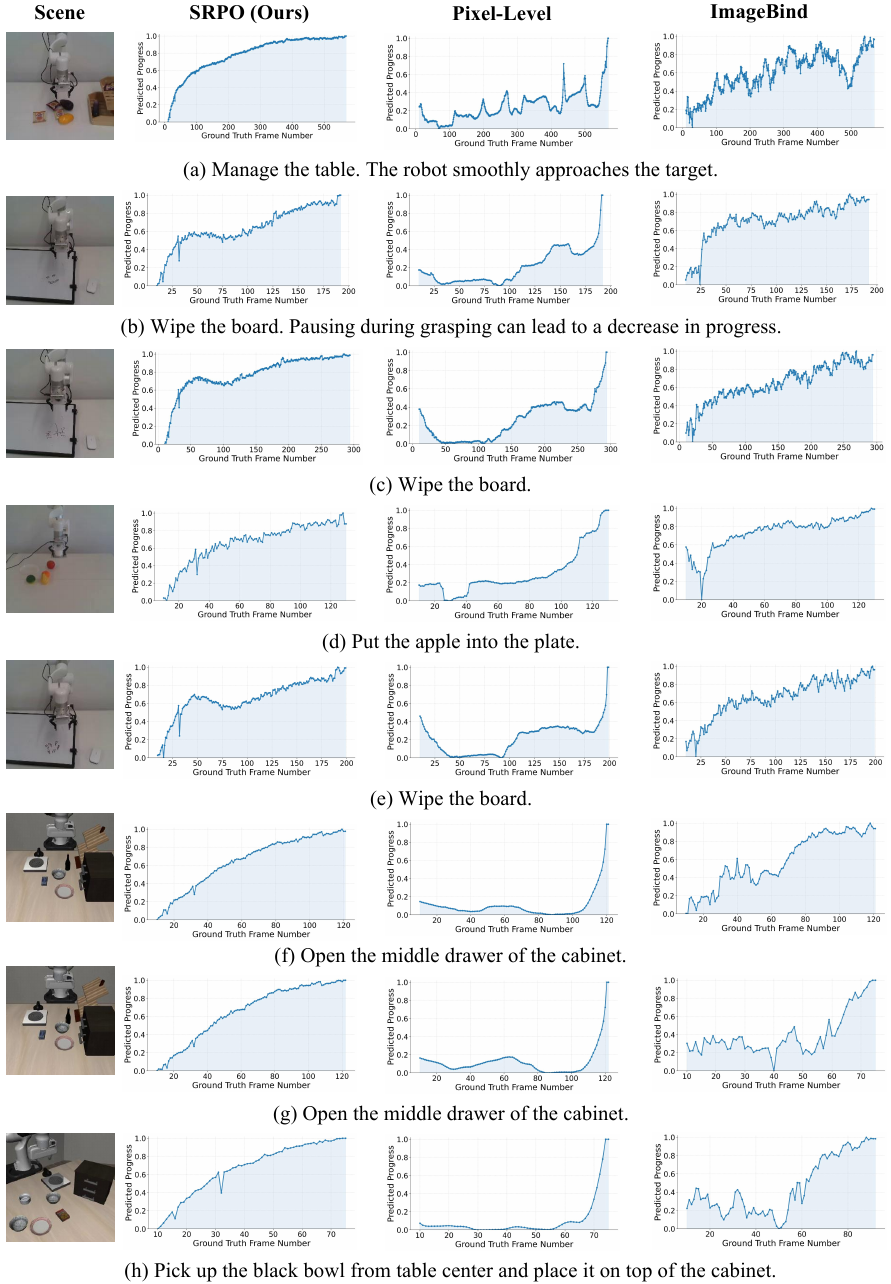}
    \caption{Reward curves of different methods on successful trajectories (Part One).}
    \label{fig:suc_curve_1}
\end{figure*}

\begin{figure*}[t]
    \centering
    \includegraphics[width=0.88\textwidth]{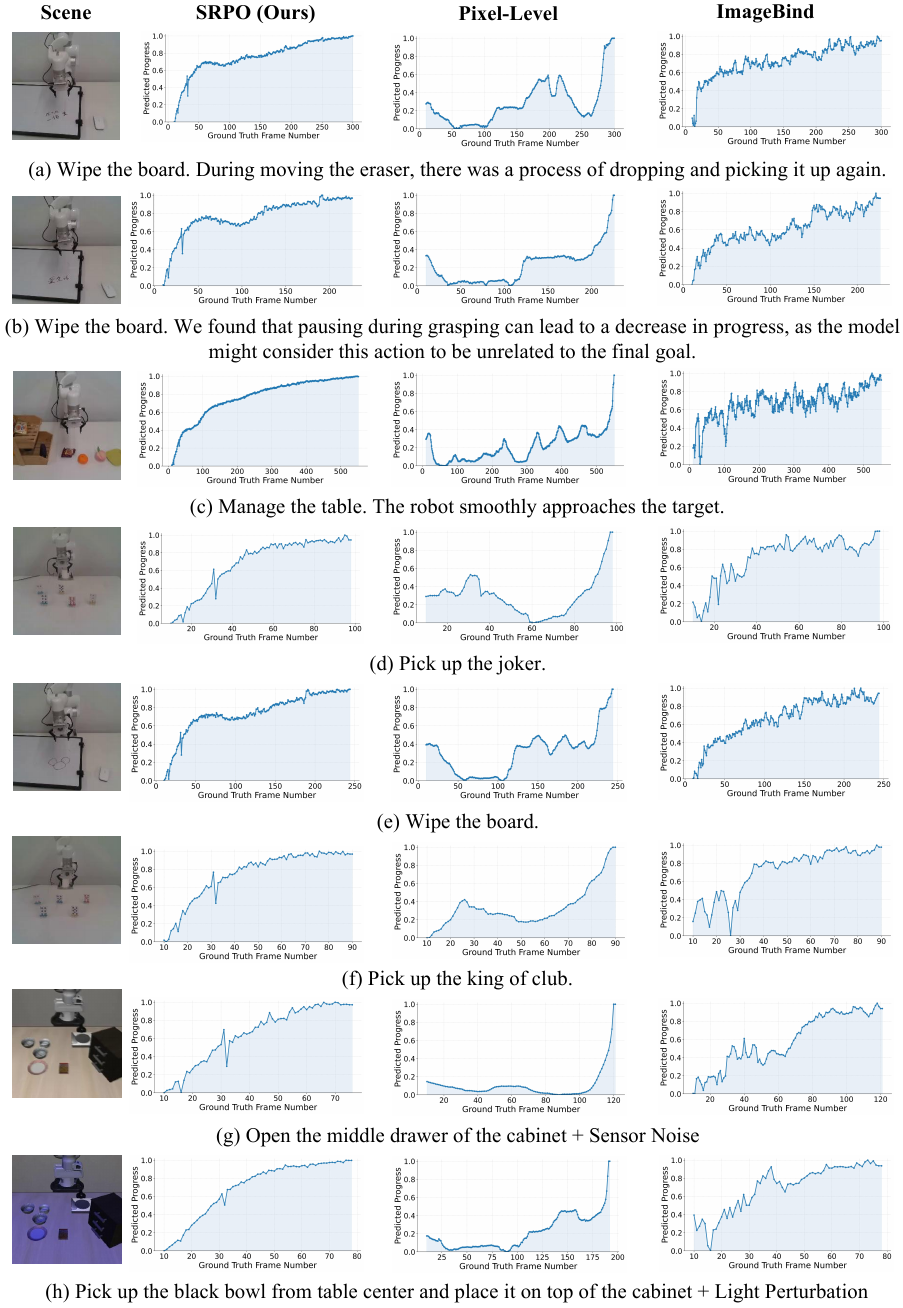}
    \caption{Reward curves of different methods on successful trajectories (Part Two).}
    \label{fig:suc_curve_2}
\end{figure*}

\begin{figure*}[t]
    \centering
    \includegraphics[width=0.85\textwidth]{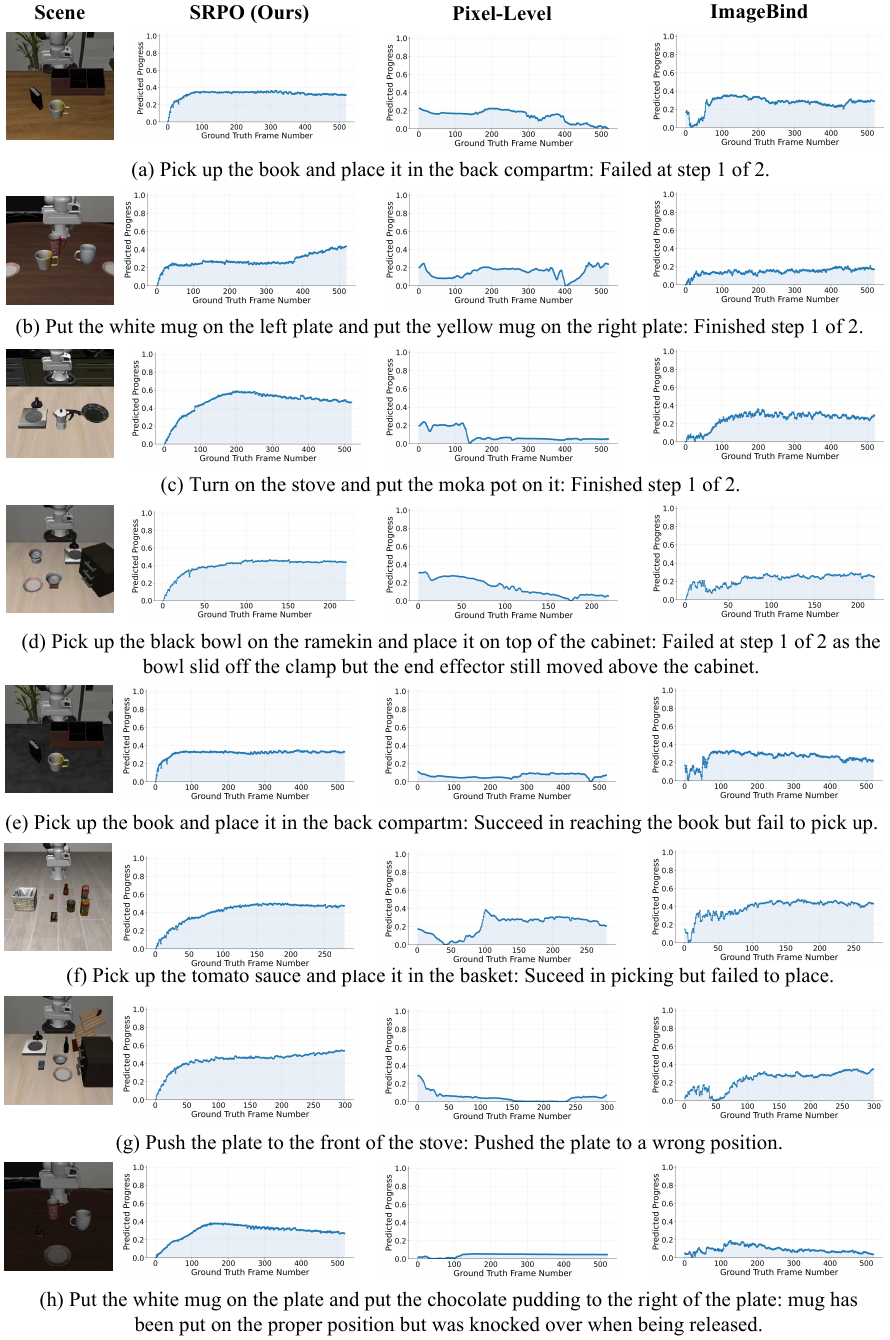}
    \caption{Reward curves of different methods on failure trajectories.}
    \label{fig:fail_curve}
\end{figure*}

\textbf{Reward Curve Analysis.} We randomly selected some successful and failure trajectories, plotting frame number-progress reward curves to compare our approach with two baseline methods. As shown in Figure~\ref{fig:suc_curve_1} \ref{fig:suc_curve_2} \ref{fig:fail_curve}, our analysis reveals that pixel-level rewards fail to properly evaluate long-horizon tasks with multiple sub-tasks and tend to exhibit sharp progress increases only in the final few frames. While general visual encoders like ImageBind can capture trajectory-level features, these features lack physical intuition, resulting in oscillatory progress rewards that are non-smooth and frequently display incorrect sudden spikes or drops—characteristics particularly unfriendly for reinforcement learning. In contrast, our reward method demonstrates more reasonable and stable progress estimation. We also provide detailed reward curves for two trajectories as visualizations in Figure~\ref{fig:detail_curve}.

\section{Ablations}
\label{app:exp_abl}

\subsection{Ablation on Self-Referential Mechanism}
\label{subsec:ablation_self_referential}

\begin{wrapfigure}{r}{0.45\linewidth}
    \centering
    \includegraphics[width=\linewidth]{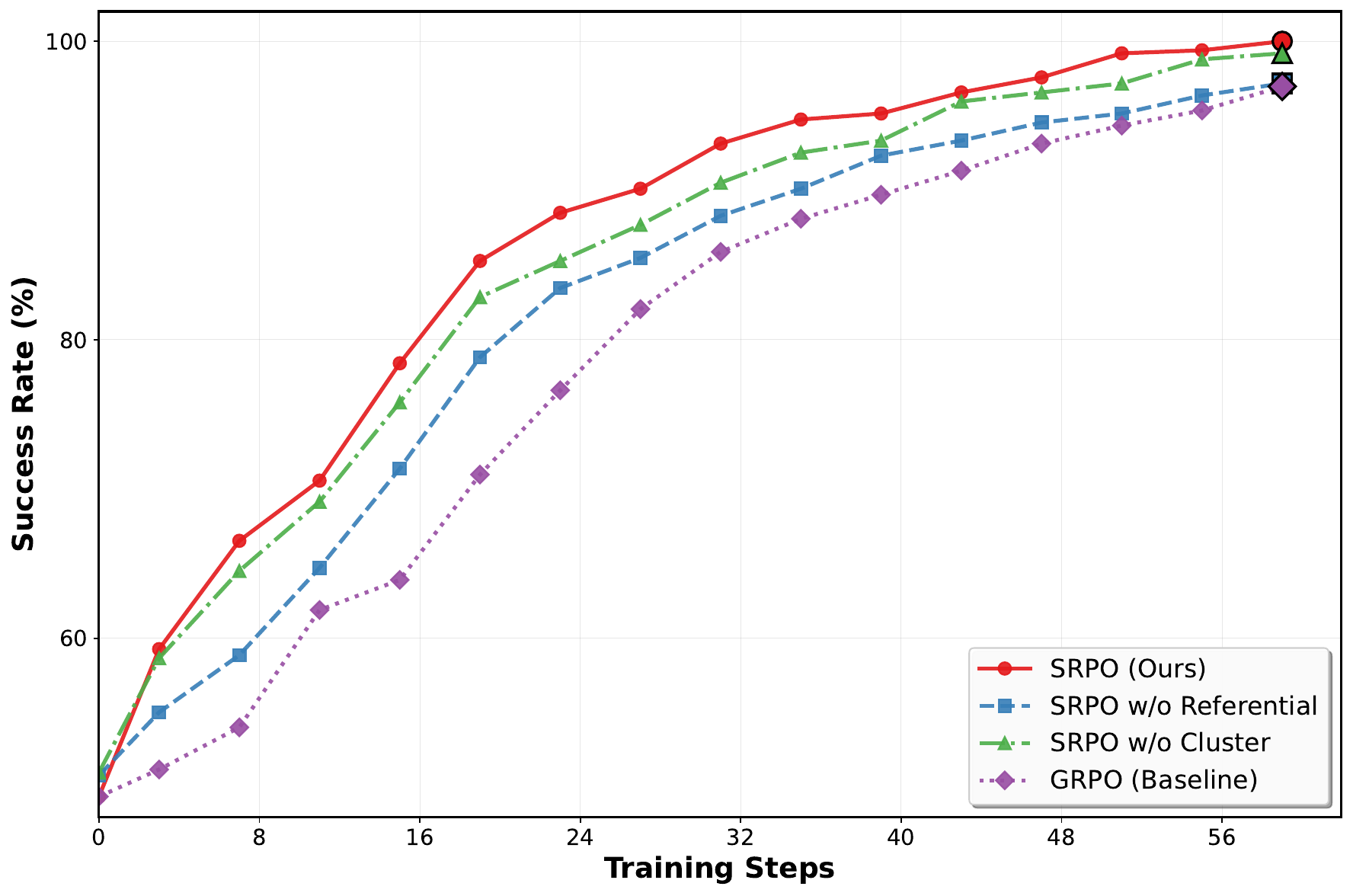}
    \caption{Ablation study on Object suite. We compare our SRPO method against its ablated variants. Removing the referential component (w/o Referential) leads to significant performance drop, while removing the clustering component (w/o Cluster) slows down convergence.}
    \label{fig:ablation}
\end{wrapfigure}

A core design of our method is its self-referential nature, which assesses progress by comparing within the current policy's own rollouts. We hypothesize that relying on a fixed set of external expert trajectories could constrain the policy's exploration and potentially lead to convergence to suboptimal local minima. To validate this, we ablate the self-referential mechanism by replacing the within-batch successful trajectories with a fixed set of 50 pre-selected expert trajectories per task for progress computation. As illustrated in Figure~\ref{fig:ablation}, the ablated variant initially trains slightly slower than our full SRPO method, yet still outperforms GRPO significantly. However, its performance plateaus in later stages, requiring nearly 1.4 times the training steps yet still yielding suboptimal results. This suggests two key insights: (1) The \textit{progress-wise} reward, even when computed from external trajectories, provides a more effective learning signal than sparse binary rewards, explaining the initial efficiency gain over GRPO, while the introduction of external information eventually allows GRPO, which requires no such information, to catch up it in performance. (2) The fixed external references ultimately limit the policy's capacity for open-ended exploration. As the policy evolves, these static trajectories fail to provide nuanced, step-by-step progress assessments for the diverse rollouts generated by the policy, leading to a performance ceiling lower than that achieved by our self-referential approach.

\subsection{Ablation on Success Clustering}
\label{subsec:ablation_clustering}

We introduce clustering of successful trajectories to compute progress rewards based on two primary motivations. First, a task can often be accomplished through multiple distinct strategies (e.g., placing object A before B, or vice versa). A failure trajectory should be compared to the nearest \textit{successful strategy} rather than an arbitrary one. Second, using the centroid of a cluster, rather than the single nearest successful trajectory, provides a more robust distance measure. Individual success trajectories might contain sub-optimal or noisy segments (e.g., a gripper moving momentarily away from an object before successfully grasping it). If a failure trajectory shares a similar initial deviation, measuring the distance solely to this specific, noisy success trajectory could yield an inaccurately high progress score. Using a cluster centroid mitigates this by representing a more prototypical and cleaner version of a success strategy.

To substantiate these points, we conduct an ablation where progress is computed using the distance to the single nearest success trajectory instead of the cluster centroid. The results are presented in Figure~\ref{fig:ablation}. We observe that this variant's initial learning efficiency is comparable to our full SRPO method. However, its performance gains diminish significantly as training progresses. We attribute this to the following: in early training stages, the repertoire of successful strategies is limited, and the number of success trajectories is small, thus diminishing the advantage of clustering. In later stages, as successful exploration diversifies and the number of success trajectories grows substantially, the ability of clustering to distill prototypical strategies and provide robust progress signals becomes crucial, leading to the increasing performance gap observed in our results.

\section{Hyperparameter Analysis}


\begin{wrapfigure}{r}{0.45\linewidth}
    \centering
    \includegraphics[width=\linewidth]{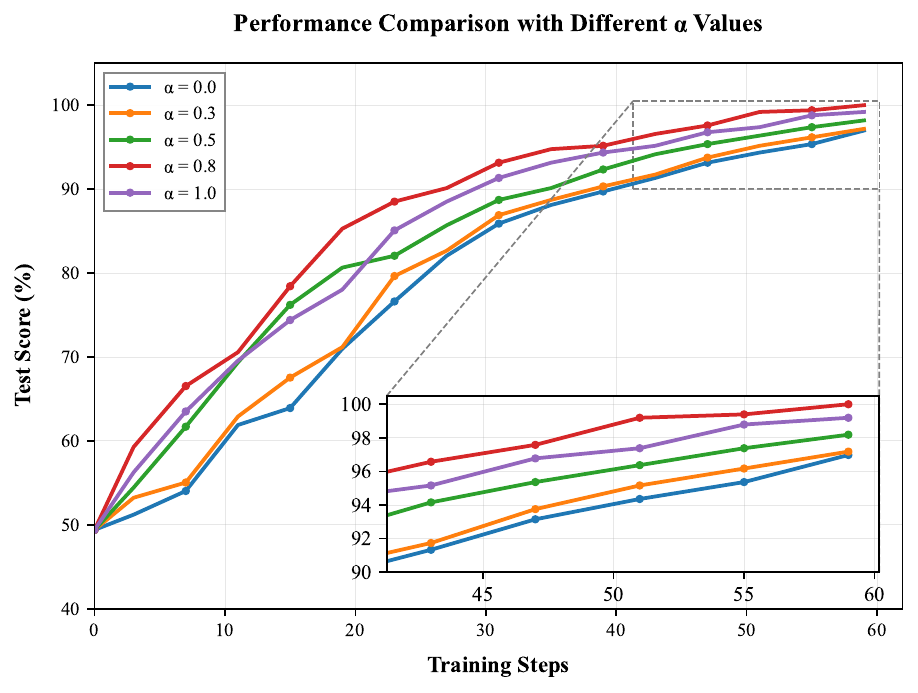}
    \caption{Performance comparison with different $\alpha$ values in the reward function. The results demonstrate that $\alpha = 0.8$ achieves the best performance, followed by $\alpha = 1.0$, $\alpha = 0.5$, $\alpha = 0.3$, and $\alpha = 0$ in descending order. This validates the importance of balancing progress awareness with outcome correctness in our reward design.}
    \label{fig:hyperparameter_analysis}
\end{wrapfigure}

In Equation \eqref{math:end}, we employ an activation function to map the reward trajectory into the range (0, 1). In our implementation, we use a sigmoid function and precede it with a scaling coefficient $\alpha$ to control the trade-off between progress awareness and outcome correctness in our reward function. We now conduct a comprehensive analysis of this hyperparameter $\alpha$. The reward design provides full credit (1.0) for correct answers and scales the progress reward by $\alpha$ otherwise. We evaluate five different values of $\alpha$: 0, 0.3, 0.5, 0.8, and 1.0.

The experimental results reveal a clear performance hierarchy: $0 < 0.3 < 0.5 < 1.0 < 0.8$. This pattern provides several important insights:

\textbf{$\alpha = 0$ (no progress reward)}: Performs worst, confirming that purely outcome-based rewards are insufficient for complex tasks requiring sequential reasoning.

$\alpha = 0.3$ and $0.5$: Show gradual improvement, indicating that even small progress rewards enhance learning efficiency.
    
\textbf{$\alpha = 1.0$ (equal weighting)}: Performs better than lower values but suboptimally, suggesting that over-emphasizing progress may distract from the final objective.
    
\textbf{$\alpha = 0.8$ (optimal)}: Achieves the best performance, demonstrating that a strong but not exclusive focus on progress rewards (80\% of maximum) provides the ideal balance for guiding the learning process while maintaining focus on task completion.

This analysis validates our reward design principle: properly weighted progress awareness ($\alpha = 0.8$) significantly outperforms both purely outcome-based rewards ($\alpha = 0$) and excessively progress-focused rewards ($\alpha = 1.0$). The optimal $\alpha$ value enables the agent to benefit from intermediate guidance while remaining oriented toward the ultimate task objective.

\section{What if we use a pixel-level world model for reward shaping?}
\label{pixel-level-cosmos}

Another promising avenue for reward shaping involves leveraging pixel-level world models to generate reference trajectories based on language instructions or action priors. Since these priors are already inputs to the policy or RL algorithm, this approach does not strictly constitute introducing external information. In this section, we use Cosmos-Predict2 \citep{ali2025world} as a case study to elucidate why this paradigm, while intuitive, often falls short in practice.

\begin{figure*}[t]
    \centering
    \includegraphics[width=0.85\textwidth]{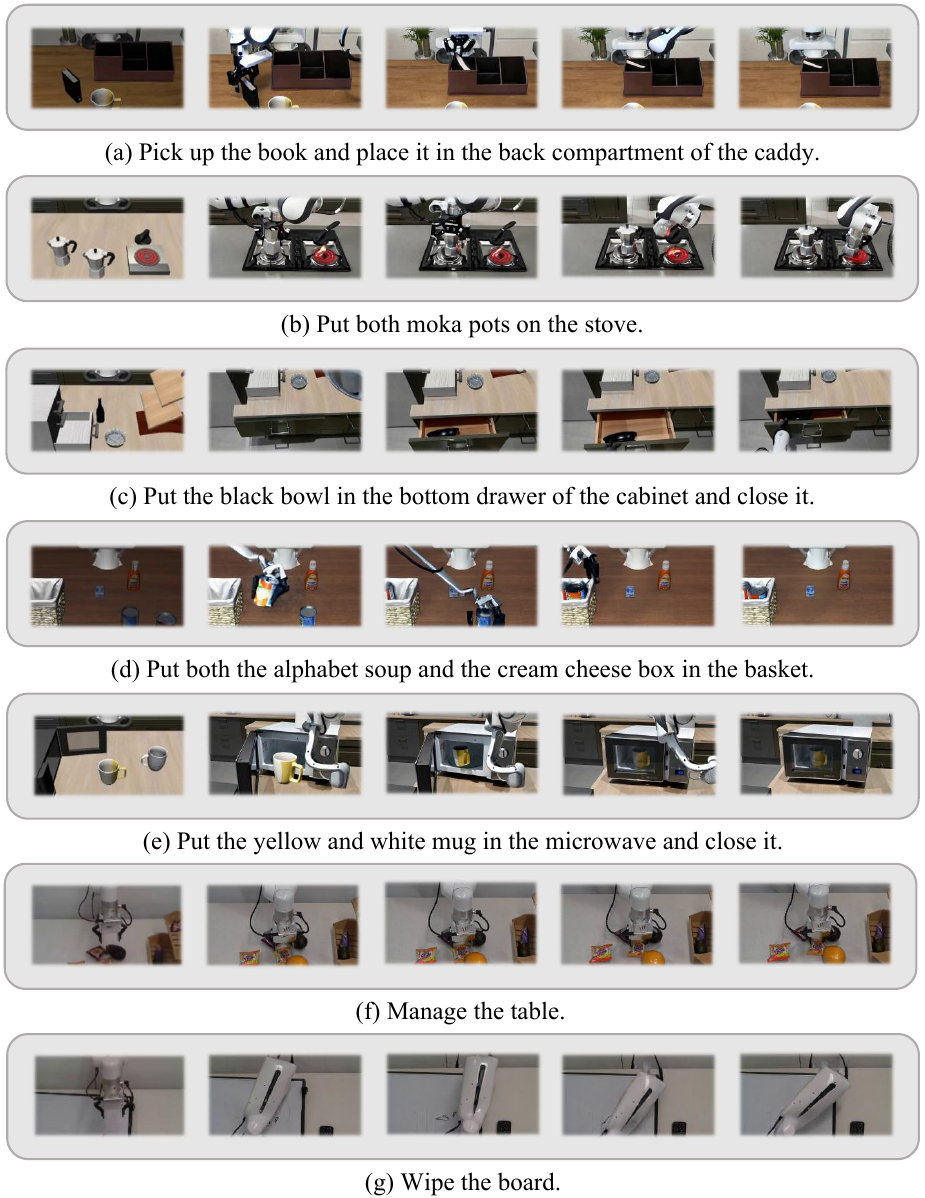}
    \caption{Trajectories generated by Cosmos-Predict2 \citep{ali2025world}.}
    \label{fig:cosmos_rollout}
\end{figure*}

\textbf{Zero-Shot Generation Yields Unsatisfactory Results.} We experimented with the large-scale Cosmos-Predict2-14B model in a zero-shot setting on several tasks from the LIBERO benchmark. Using the language instruction as conditioning signals, the model was tasked with generating a reference video trajectory. As shown in Figure~\ref{fig:cosmos_rollout}, the generated videos suffer from poor scene consistency. Furthermore, although task-specific supervised fine-tuning (SFT) may mitigate these inconsistencies and improve generation quality, this high-cost approach, which relies heavily on extensive expert demonstrations, is clearly inferior to the reward modeling paradigm proposed by SRPO, which is based on a latent world representation and proves to be more cost-effective and generalizable.

\section{Training Details}
\label{libero_train_detail}

\subsection{Supervised Fine-Tuning (SFT) Stage}
Our one-shot SFT phase builds upon the OpenVLA (with Action Chunking and Parallel Decoding) checkpoint. The training was conducted on $8 \times \text{A100}$ GPUs. Key training configurations are as follows:

\begin{itemize}
    \item \textbf{Optimizer:} AdamW
    \item \textbf{Learning Rate:} $5 \times 10^{-4}$
    \item \textbf{Batch Size:} 8
    \item \textbf{Maximum Training Steps:} 150,005
    \item \textbf{Learning Rate Decay:} Applied after 100,000 steps
    \item \textbf{LoRA Rank:} 32
    \item \textbf{Image Augmentation:} Enabled
    \item \textbf{Input Modalities:} Text instruction and single image observation
    \item \textbf{Action Chunking:} 8 action chunks
\end{itemize}

\subsection{SRPO Post-Training Stage}
The SRPO reinforcement learning phase is initialized from the one-shot SFT model. Key hyperparameters include:

\begin{itemize}
    \item \textbf{Algorithm:} SRPO (based on GRPO advantage estimation)
    \item \textbf{Learning Rate:} $5 \times 10^{-6}$
    \item \textbf{Samples Per Group:} 8
    \item \textbf{Batch Size:} 64 ($\times 8$, training), 496 (validation)
    \item \textbf{Mini-batch Size:} 128
    \item \textbf{Progress Reward Weight:} 0.8
    \item \textbf{Number of Trials per Task:} 50
    \item \textbf{Action Configuration:} 7 action tokens, 8 action chunks
    \item \textbf{Context Length:} 512 tokens (prompt), 128 tokens (response)
    \item \textbf{Trajectory Mini-batch Size:} 16
    \item \textbf{Log Probability Batch Size:} 8 (for both rollout and reference model)
    \item \textbf{FP16 Inference:} Enabled for the video embedding model
    \item \textbf{Model Offloading:} Enabled to optimize GPU memory usage
\end{itemize}

\subsection{Model Details}
\label{app:model_details}

This modified architecture retains the Llama 2 backbone for generating discrete action tokens, in contrast to the continuous action heads employed in OpenVLA-OFT. While this design choice may potentially sacrifice some level of action precision compared to continuous output methods, it provides the crucial advantage of enabling direct access to action log-probabilities essential for policy gradient methods in reinforcement learning.

\section{Real-world Experiment Details}

\subsection{Detailed Setup}
\label{exp:real-world-details}
In our physical robot experiments, due to safety concerns associated with online exploration and the significant time cost of manual resets, we adopted an offline reinforcement learning (RL) paradigm. Specifically, our technical approach integrates the Advantage-Weighted Regression (AWR) strategy \citep{peng2019advantage} with SRPO's self-referential progress-wise reward and advantage mechanism. We first collect demonstration data and store it in a trajectory buffer. The expected cumulative reward (or value) $R_{i,t}$ for the $i$-th trajectory at step $t$ is computed as in \eqref{math:begin}--\eqref{math:end}. We then define the incremental progress at step $t$ as $D_{i,t} = R_{i,t} - R_{i,t-1}$. Following the SRPO framework, the advantage function is calculated as:
\begin{align}
    A_{i,t} = \frac{D_{i,t} - \mu}{\sigma},
\end{align}
where $\mu$ and $\sigma$ are the mean and standard deviation of $D_{i,t}$ computed across all trajectories. Our real-world task policy is trained from scratch via RL starting from pretrained weights, with the baseline for comparison being the SFT model.

The pick-and-place tasks include \textit{put apple into the plate} and \textit{put pear into the plate}. To test robust object identification, we place multiple fruits on the table initially, requiring the model to distinguish target objects from distractors. During task execution, we randomly swap positions between target and distractor objects to assess the model's capacity for rapid adaptation.

\begin{wraptable}{r}{0.5\linewidth}
\centering
\small
\caption{Progress Reward Benchmark results on real-robot datasets. Our method maintains strong performance across all tasks and metrics, demonstrating robust generalization to diverse real-world manipulation tasks.}
\label{tab:real_world_benchmark}
\begin{tabular}{lccccc}
\toprule
\textbf{Task} & \textbf{SC} & \textbf{Mono} & \textbf{MMD} & \textbf{JS} & \textbf{SMD} \\
\midrule
\textbf{Put Apple} & 0.987 & 0.975 & 0.589 & 0.562 & 165.3 \\
\textbf{Put Pear} & 0.991 & 0.982 & 0.601 & 0.578 & 172.8 \\
\textbf{Fold Towel} & 0.984 & 0.968 & 0.572 & 0.549 & 158.6 \\
\textbf{Wipe Board} & 0.993 & 0.986 & 0.624 & 0.591 & 181.2 \\
\textbf{Select Poker} & 0.989 & 0.979 & 0.595 & 0.569 & 169.5 \\
\midrule
\textbf{Average} & 0.989 & 0.978 & 0.596 & 0.570 & 169.5 \\
\bottomrule
\end{tabular}
\end{wraptable}

To specifically demonstrate the advantages of dense progress-aware rewards in complex manipulation, we incorporate \textit{folding towels} involving deformable object manipulation and \textit{cleaning whiteboard} requiring coordinated surface contact. Furthermore, to evaluate semantic understanding capabilities, we include the \textit{select poker} task where the model must identify  the specified card from these five playing cards: Joker, Jack of Spades, King of Club, Jack of Spades, 10 of Spades.

Our real-world experimental results demonstrate significant performance improvements across all five manipulation tasks when applying progress-aware value weighting. As shown in Figure~\ref{fig:real_world_results}, both VLA policy backbones exhibit substantial gains over their SFT counterparts.

\subsection{Rewarding Validation}
\label{exp:real-world-reward}

To quantitatively validate the generalization capability of our progress reward model to real-world scenarios, we evaluate our pre-trained progress reward function on five real-robot manipulation tasks, each containing 30 successful trajectories and 20 failure trajectories. The comprehensive Progress Reward Benchmark results across all tasks demonstrate consistent performance in real-world settings:

As shown in Table~\ref{tab:real_world_benchmark}, the consistently high Progress Reward Quality scores across all five real-robot tasks confirm that our latent progress reward modeling effectively captures task progression dynamics in real-world robotic manipulation. With near-perfect Spearman correlation and monotonicity, coupled with strong distribution separation in MMD and SMD, our method demonstrates robust generalization despite domain shifts from simulation to reality.

\subsection{Case Study}
\label{exp:case-study}

\begin{figure*}[t]
    \centering
    \includegraphics[width=0.86\textwidth]{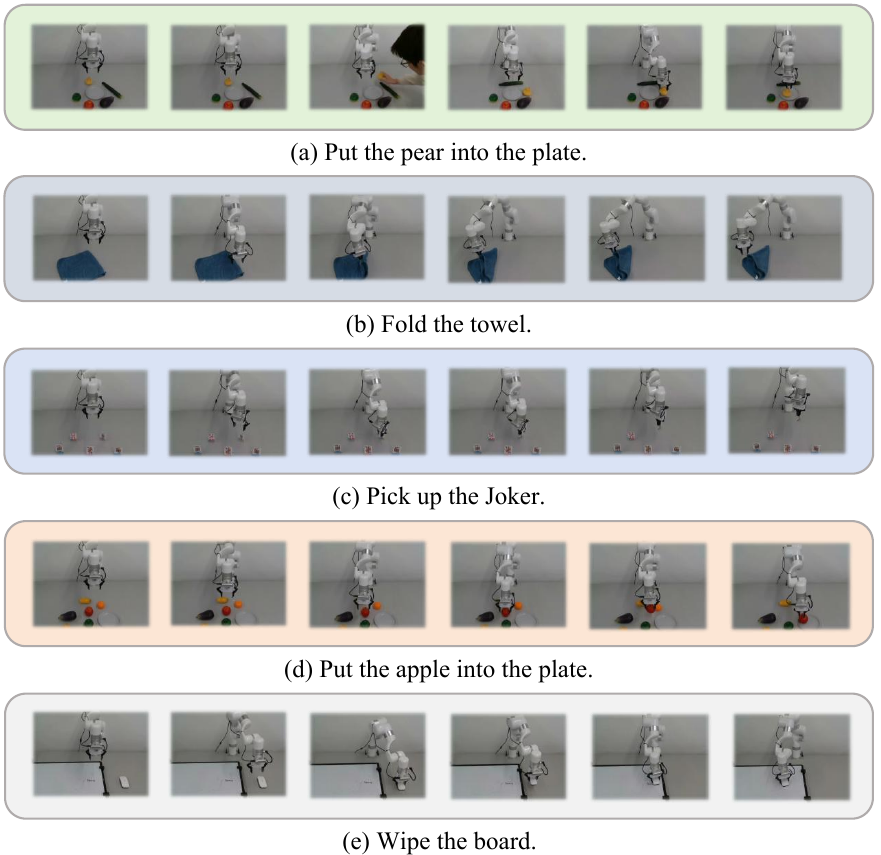}
    \caption{Success cases of the real-world experiment.}
    \label{fig:suc_real_exp}
\end{figure*}

As shown in Figure~\ref{fig:suc_real_exp}, our real-world case studies demonstrate the robustness and generalization of our approach across diverse scenarios: in \textit{put the pear into the plate}, the policy dynamically replans when the target is moved mid-execution; in \textit{fold the towel}, it reliably handles deformable object manipulation; and in \textit{pick up the Joker card}, it maintains precise semantic understanding to identify the target among multiple options. These results validate the adaptability and task-aware capability of our method in challenging real environments.

\end{document}